\documentclass[10pt,twocolumn,letterpaper]{article}
\usepackage[pagenumbers]{cvpr} % To force page numbers, e.g. for an arXiv version

\usepackage[dvipsnames]{xcolor}

\newcommand{\parag}[1]{\smallskip\noindent\textbf{#1}\enspace}

\definecolor{cvprblue}{rgb}{0.21,0.49,0.74}
\usepackage[pagebackref,breaklinks,colorlinks,citecolor=cvprblue]{hyperref}
\usepackage{comment}
	
\usepackage{color, colortbl}

\usepackage{subcaption} 
\usepackage{pifont}
\usepackage{array}

\usepackage{tikz}
\usepackage{pgfplots}\pgfplotsset{compat=1.9}
\usetikzlibrary{pgfplots.groupplots}

\usepackage{cuted}
\usepackage{makecell}
\newcommand{\xmark}{\ding{55}}%

\usepackage{dsfont}

\newif\ifshowedits

\newcommand{\addeditor}[3]{%
  \definecolor{#1color}{rgb}{#3}
  \expandafter\newcommand\csname#1\endcsname[1]{
  \ifshowedits
    {\color{#1color}##1}
  \else
    {##1}
  \fi
  }%
  \expandafter\newcommand\csname #1c\endcsname[1]{
  \ifshowedits
    {\color{#1color}{\bf[#2: ##1]}}
  \else
    {}
  \fi
  }%
}

\newcommand{\newtvar}[1]{
  \expandafter\newcommand\csname #1\endcsname{\text{#1}}
}

\newcommand{\dinocorr}{A^{\xi}}
\newcommand{\clipcorr}{A^{\phi}}

\newcommand{\clipfeat}{\phi^L(X)} %\mathbf{F}_{\text{C{lip}}}}
\newcommand{\clipinterfeat}{\phi^l(X)} %\mathbf{F}_{\text{C{lip}}}}
\newcommand{\clippatch}[1]{F_{#1}}

\newcommand{\promptstyle}[2]{\footnotesize{\textbf{\texttt{\color{#1}#2}}}}

\newcommand{\addresult}[3]{%
\begin{minipage}[b]{#1}
     \centering
     \includegraphics[width=\textwidth, height=\resultsheight]{figures/images/#2_#3.png}
\end{minipage}}
\newcommand{\addresultjpeg}[2]{%
\begin{minipage}[b]{#1}
     \centering
     \includegraphics[width=\textwidth, height=\resultsheight]{figures/images/#2.jpg}
\end{minipage}}

\newcommand{\addcocoablresult}[3]{%
\begin{minipage}[b]{#1}
     \centering
     \includegraphics[width=\textwidth, height=\ablcocoheight]{figures/images/#2_#3.png}
\end{minipage}}
\newcommand{\addcocoablresultjpeg}[2]{%
\begin{minipage}[b]{#1}
     \centering
     \includegraphics[width=\textwidth, height=\ablcocoheight]{figures/images/#2.jpg}
\end{minipage}}
\definecolor{cow}{RGB}{64, 128, 0}
\definecolor{sheep}{RGB}{128, 64, 0}
\definecolor{train}{RGB}{128, 192, 0}
\definecolor{dog}{RGB}{64, 0, 128}
\definecolor{sofa}{RGB}{0, 192, 0}
\definecolor{surf}{RGB}{0, 224, 128}
\definecolor{umbrella}{RGB}{0, 160, 0}
\definecolor{elephant}{RGB}{128, 128, 0}
\definecolor{bear}{RGB}{200, 0, 32}
\definecolor{zebra}{RGB}{128, 32, 0}
\definecolor{remote}{RGB}{128, 192, 192}
\definecolor{fork}{RGB}{64, 0, 160}
\definecolor{bowl}{RGB}{192, 0, 32}
\definecolor{cellphone}{RGB}{192, 160, 128}
\definecolor{sink}{RGB}{128, 64, 128}
\definecolor{hair}{RGB}{64, 128, 224}
\definecolor{bird_pascal}{RGB}{128, 128, 0}
\definecolor{aeroplane}{RGB}{190, 0, 0}
\definecolor{bottle}{RGB}{128, 0, 128}
\definecolor{tv/monitor}{RGB}{0, 64, 128}
\definecolor{bottle}{RGB}{200, 0, 200}
\definecolor{chair}{RGB}{192, 0, 0}
\definecolor{plant}{RGB}{192, 32, 128}
\definecolor{bench}{RGB}{0, 160, 192}
\definecolor{orange}{RGB}{64, 224, 0}
\definecolor{vase}{RGB}{192, 128, 224}
\definecolor{scissors}{RGB}{255, 140, 31}
\definecolor{thread}{RGB}{10, 206, 250}
\definecolor{apple}{RGB}{0, 128, 128}
\definecolor{darkgreen}{RGB}{0, 200, 20}
\definecolor{dumplings}{RGB}{128, 0, 128}
\definecolor{bus}{RGB}{0, 128, 128}
\definecolor{boat}{RGB}{0, 0, 128}
\definecolor{plant_pascal}{RGB}{0, 64, 0}
\definecolor{bike}{RGB}{0, 128, 0}
\definecolor{horse}{RGB}{128, 128, 224}
\definecolor{pizza}{RGB}{0, 140, 220}
\definecolor{spoon}{RGB}{128, 128, 64}
\definecolor{cake}{RGB}{192, 224, 0}
\definecolor{bowl}{RGB}{192, 0, 32}
\definecolor{microwave}{RGB}{128, 192, 0}
\definecolor{refrigerator}{RGB}{64, 128, 96}
\definecolor{sandwich}{RGB}{0, 128, 32}
\definecolor{cup}{RGB}{128, 128, 128}
\definecolor{knife}{RGB}{128, 224, 128}
\definecolor{cat}{RGB}{128, 0, 192}
\definecolor{carrot}{RGB}{80, 80, 150}
\definecolor{diningtable}{RGB}{192, 128, 0}
\definecolor{nemo}{RGB}{255, 80, 31}
\definecolor{dory}{RGB}{10, 170, 250}
\definecolor{greyeleph}{RGB}{100, 170, 150}
\definecolor{pinkeleph}{RGB}{243, 10, 198}
\definecolor{pasteis}{RGB}{0, 70, 150}
\definecolor{coffee}{RGB}{143, 107, 8}
\definecolor{scrabble}{RGB}{10, 206, 250}
\definecolor{pencil}{RGB}{255, 140, 31}
\definecolor{rubber stamp}{RGB}{55, 240, 31}
\definecolor{paper clip}{RGB}{200, 0, 25}

\definecolor{background_coco}{RGB}{0, 0, 0}
\definecolor{person_coco}{RGB}{0, 192, 64}
\definecolor{bicycle_coco}{RGB}{0, 192, 64}
\definecolor{car_coco}{RGB}{0, 64, 96}
\definecolor{motorcycle_coco}{RGB}{128, 192, 192}
\definecolor{airplane_coco}{RGB}{0, 64, 64}
\definecolor{bus_coco}{RGB}{0, 192, 224}
\definecolor{train_coco}{RGB}{0, 192, 192}
\definecolor{truck_coco}{RGB}{128, 192, 64}
\definecolor{boat_coco}{RGB}{0, 192, 96}
\definecolor{traffic light_coco}{RGB}{128, 192, 64}
\definecolor{fire hydrant_coco}{RGB}{128, 32, 192}
\definecolor{stop sign_coco}{RGB}{0, 0, 224}
\definecolor{parking meter_coco}{RGB}{0, 0, 64}
\definecolor{bench_coco}{RGB}{0, 160, 192}
\definecolor{bird_coco}{RGB}{128, 0, 96}
\definecolor{cat_coco}{RGB}{128, 0, 192}
\definecolor{dog_coco}{RGB}{0, 32, 192}
\definecolor{horse_coco}{RGB}{128, 128, 224}
\definecolor{sheep_coco}{RGB}{0, 0, 192}
\definecolor{cow_coco}{RGB}{128, 160, 192}
\definecolor{elephant_coco}{RGB}{128, 128, 0}
\definecolor{bear_coco}{RGB}{128, 0, 32}
\definecolor{zebra_coco}{RGB}{128, 32, 0}
\definecolor{giraffe_coco}{RGB}{128, 0, 128}
\definecolor{backpack_coco}{RGB}{64, 128, 32}
\definecolor{umbrella_coco}{RGB}{0, 160, 0}
\definecolor{handbag_coco}{RGB}{0, 0, 0}
\definecolor{tie_coco}{RGB}{192, 128, 160}
\definecolor{suitcase_coco}{RGB}{0, 32, 0}
\definecolor{frisbee_coco}{RGB}{0, 128, 128}
\definecolor{skis_coco}{RGB}{64, 128, 160}
\definecolor{snowboard_coco}{RGB}{128, 160, 0}
\definecolor{sports ball_coco}{RGB}{0, 128, 0}
\definecolor{kite_coco}{RGB}{192, 128, 32}
\definecolor{baseball bat_coco}{RGB}{128, 96, 128}
\definecolor{baseball glove_coco}{RGB}{0, 0, 128}
\definecolor{skateboard_coco}{RGB}{64, 0, 32}
\definecolor{surfboard_coco}{RGB}{0, 224, 128}
\definecolor{tennis racket_coco}{RGB}{128, 0, 0}
\definecolor{bottle_coco}{RGB}{192, 0, 160}
\definecolor{wine glass_coco}{RGB}{0, 96, 128}
\definecolor{cup_coco}{RGB}{128, 128, 128}
\definecolor{fork_coco}{RGB}{64, 0, 160}
\definecolor{knife_coco}{RGB}{128, 224, 128}
\definecolor{spoon_coco}{RGB}{128, 128, 64}
\definecolor{bowl_coco}{RGB}{192, 0, 32}
\definecolor{banana_coco}{RGB}{128, 96, 0}
\definecolor{apple_coco}{RGB}{128, 0, 192}
\definecolor{sandwich_coco}{RGB}{0, 128, 32}
\definecolor{orange_coco}{RGB}{64, 224, 0}
\definecolor{broccoli_coco}{RGB}{0, 0, 64}
\definecolor{carrot_coco}{RGB}{128, 128, 160}
\definecolor{hot dog_coco}{RGB}{64, 96, 0}
\definecolor{pizza_coco}{RGB}{0, 128, 192}
\definecolor{donut_coco}{RGB}{0, 128, 160}
\definecolor{cake_coco}{RGB}{192, 224, 0}
\definecolor{chair_coco}{RGB}{0, 128, 64}
\definecolor{couch_coco}{RGB}{128, 128, 32}
\definecolor{potted plant_coco}{RGB}{192, 32, 128}
\definecolor{bed_coco}{RGB}{0, 64, 192}
\definecolor{dining table_coco}{RGB}{0, 0, 32}
\definecolor{toilet_coco}{RGB}{64, 160, 128}
\definecolor{tv_coco}{RGB}{128, 64, 64}
\definecolor{laptop_coco}{RGB}{128, 0, 160}
\definecolor{mouse_coco}{RGB}{64, 32, 128}
\definecolor{remote_coco}{RGB}{128, 192, 192}
\definecolor{keyboard_coco}{RGB}{0, 0, 160}
\definecolor{cell phone_coco}{RGB}{192, 160, 128}
\definecolor{microwave_coco}{RGB}{128, 192, 0}
\definecolor{oven_coco}{RGB}{128, 0, 96}
\definecolor{toaster_coco}{RGB}{192, 32, 0}
\definecolor{sink_coco}{RGB}{128, 64, 128}
\definecolor{refrigerator_coco}{RGB}{64, 128, 96}
\definecolor{book_coco}{RGB}{64, 160, 0}
\definecolor{clock_coco}{RGB}{0, 64, 0}
\definecolor{vase_coco}{RGB}{192, 128, 224}
\definecolor{scissors_coco}{RGB}{64, 32, 0}
\definecolor{teddy bear_coco}{RGB}{0, 192, 128}
\definecolor{hair drier_coco}{RGB}{64, 128, 224}
\definecolor{toothbrush_coco}{RGB}{192, 160, 0}

\definecolor{road_city}{RGB}{128, 64, 128}
\definecolor{sidewalk_city}{RGB}{244, 35, 232}
\definecolor{building_city}{RGB}{70, 70, 70}
\definecolor{wall_city}{RGB}{102, 102, 156}
\definecolor{fence_city}{RGB}{190, 153, 153}
\definecolor{pole_city}{RGB}{153, 153, 153}
\definecolor{traffic_light_city}{RGB}{250, 170, 30}
\definecolor{traffic_sign_city}{RGB}{220, 220, 0}
\definecolor{vegetation_city}{RGB}{107, 142, 35}
\definecolor{terrain_city}{RGB}{152, 251, 152}
\definecolor{sky_city}{RGB}{70, 130, 180}
\definecolor{person_city}{RGB}{220, 20, 60}
\definecolor{rider_city}{RGB}{255, 0, 0}
\definecolor{car_city}{RGB}{0, 0, 142}
\definecolor{truck_city}{RGB}{0, 0, 70}
\definecolor{bus_city}{RGB}{0, 60, 100}
\definecolor{train_city}{RGB}{0, 80, 100}
\definecolor{motorcycle_city}{RGB}{0, 0, 230}
\definecolor{bicycle_city}{RGB}{119, 11, 32}

\definecolor{aeroplane_voc}{RGB}{128, 0, 0}
\definecolor{bicycle_voc}{RGB}{0, 128, 0}
\definecolor{bird_voc}{RGB}{128, 128, 0}
\definecolor{boat_voc}{RGB}{0, 0, 128}
\definecolor{bottle_voc}{RGB}{128, 0, 128}
\definecolor{bus_voc}{RGB}{0, 128, 128}
\definecolor{car_voc}{RGB}{128, 128, 128}
\definecolor{cat_voc}{RGB}{64, 0, 0}
\definecolor{chair_voc}{RGB}{192, 0, 0}
\definecolor{dog_voc}{RGB}{64, 0, 128}
\definecolor{horse_voc}{RGB}{192, 0, 128}
\definecolor{motorbike_voc}{RGB}{64, 128, 128}
\definecolor{person_voc}{RGB}{192, 128, 128}
\definecolor{pottedplant_voc}{RGB}{0, 64, 0}
\definecolor{sheep_voc}{RGB}{128, 64, 0}
\definecolor{sofa_voc}{RGB}{0, 192, 0}
\definecolor{train_voc}{RGB}{128, 192, 0}
\definecolor{tvmonitor_voc}{RGB}{0, 64, 128}

\definecolor{wall_ade}{RGB}{120, 120, 120}
\definecolor{building_ade}{RGB}{180, 120, 120}
\definecolor{sky_ade}{RGB}{6, 230, 230}
\definecolor{floor_ade}{RGB}{80, 50, 50}
\definecolor{tree_ade}{RGB}{4, 200, 3}
\definecolor{ceiling_ade}{RGB}{120, 120, 80}
\definecolor{road_ade}{RGB}{140, 140, 140}
\definecolor{bed _ade}{RGB}{204, 5, 255}
\definecolor{windowpane_ade}{RGB}{230, 230, 230}
\definecolor{grass_ade}{RGB}{4, 250, 7}
\definecolor{cabinet_ade}{RGB}{224, 5, 255}
\definecolor{sidewalk_ade}{RGB}{235, 255, 7}
\definecolor{person_ade}{RGB}{150, 5, 61}
\definecolor{earth_ade}{RGB}{120, 120, 70}
\definecolor{door_ade}{RGB}{8, 255, 51}
\definecolor{table_ade}{RGB}{255, 6, 82}
\definecolor{mountain_ade}{RGB}{143, 255, 140}
\definecolor{plant_ade}{RGB}{204, 255, 4}
\definecolor{curtain_ade}{RGB}{255, 51, 7}
\definecolor{chair_ade}{RGB}{204, 70, 3}
\definecolor{car_ade}{RGB}{0, 102, 200}
\definecolor{water_ade}{RGB}{61, 230, 250}
\definecolor{painting_ade}{RGB}{255, 6, 51}
\definecolor{sofa_ade}{RGB}{11, 102, 255}
\definecolor{shelf_ade}{RGB}{255, 7, 71}
\definecolor{house_ade}{RGB}{255, 9, 224}
\definecolor{sea_ade}{RGB}{9, 7, 230}
\definecolor{mirror_ade}{RGB}{220, 220, 220}
\definecolor{rug_ade}{RGB}{255, 9, 92}
\definecolor{field_ade}{RGB}{112, 9, 255}
\definecolor{armchair_ade}{RGB}{8, 255, 214}
\definecolor{seat_ade}{RGB}{7, 255, 224}
\definecolor{fence_ade}{RGB}{255, 184, 6}
\definecolor{desk_ade}{RGB}{10, 255, 71}
\definecolor{rock_ade}{RGB}{255, 41, 10}
\definecolor{wardrobe_ade}{RGB}{7, 255, 255}
\definecolor{lamp_ade}{RGB}{224, 255, 8}
\definecolor{bathtub_ade}{RGB}{102, 8, 255}
\definecolor{railing_ade}{RGB}{255, 61, 6}
\definecolor{cushion_ade}{RGB}{255, 194, 7}
\definecolor{base_ade}{RGB}{255, 122, 8}
\definecolor{box_ade}{RGB}{0, 255, 20}
\definecolor{column_ade}{RGB}{255, 8, 41}
\definecolor{signboard_ade}{RGB}{255, 5, 153}
\definecolor{chest of drawers_ade}{RGB}{6, 51, 255}
\definecolor{counter_ade}{RGB}{235, 12, 255}
\definecolor{sand_ade}{RGB}{160, 150, 20}
\definecolor{sink_ade}{RGB}{0, 163, 255}
\definecolor{skyscraper_ade}{RGB}{140, 140, 140}
\definecolor{fireplace_ade}{RGB}{250, 10, 15}
\definecolor{refrigerator_ade}{RGB}{20, 255, 0}
\definecolor{grandstand_ade}{RGB}{31, 255, 0}
\definecolor{path_ade}{RGB}{255, 31, 0}
\definecolor{stairs_ade}{RGB}{255, 224, 0}
\definecolor{runway_ade}{RGB}{153, 255, 0}
\definecolor{case_ade}{RGB}{0, 0, 255}
\definecolor{pool table_ade}{RGB}{255, 71, 0}
\definecolor{pillow_ade}{RGB}{0, 235, 255}
\definecolor{screen door_ade}{RGB}{0, 173, 255}
\definecolor{stairway_ade}{RGB}{31, 0, 255}
\definecolor{river_ade}{RGB}{11, 200, 200}
\definecolor{bridge_ade}{RGB}{255, 82, 0}
\definecolor{bookcase_ade}{RGB}{0, 255, 245}
\definecolor{blind_ade}{RGB}{0, 61, 255}
\definecolor{coffee table_ade}{RGB}{0, 255, 112}
\definecolor{toilet_ade}{RGB}{0, 255, 133}
\definecolor{flower_ade}{RGB}{255, 0, 0}
\definecolor{book_ade}{RGB}{255, 163, 0}
\definecolor{hill_ade}{RGB}{255, 102, 0}
\definecolor{bench_ade}{RGB}{194, 255, 0}
\definecolor{countertop_ade}{RGB}{0, 143, 255}
\definecolor{stove_ade}{RGB}{51, 255, 0}
\definecolor{palm_ade}{RGB}{0, 82, 255}
\definecolor{kitchen island_ade}{RGB}{0, 255, 41}
\definecolor{computer_ade}{RGB}{0, 255, 173}
\definecolor{swivel chair_ade}{RGB}{10, 0, 255}
\definecolor{boat_ade}{RGB}{173, 255, 0}
\definecolor{bar_ade}{RGB}{0, 255, 153}
\definecolor{arcade machine_ade}{RGB}{255, 92, 0}
\definecolor{hovel_ade}{RGB}{255, 0, 255}
\definecolor{bus_ade}{RGB}{255, 0, 245}
\definecolor{towel_ade}{RGB}{255, 0, 102}
\definecolor{light_ade}{RGB}{255, 173, 0}
\definecolor{truck_ade}{RGB}{255, 0, 20}
\definecolor{tower_ade}{RGB}{255, 184, 184}
\definecolor{chandelier_ade}{RGB}{0, 31, 255}
\definecolor{awning_ade}{RGB}{0, 255, 61}
\definecolor{streetlight_ade}{RGB}{0, 71, 255}
\definecolor{booth_ade}{RGB}{255, 0, 204}
\definecolor{television receiver_ade}{RGB}{0, 255, 194}
\definecolor{airplane_ade}{RGB}{0, 255, 82}
\definecolor{dirt track_ade}{RGB}{0, 10, 255}
\definecolor{apparel_ade}{RGB}{0, 112, 255}
\definecolor{pole_ade}{RGB}{51, 0, 255}
\definecolor{land_ade}{RGB}{0, 194, 255}
\definecolor{bannister_ade}{RGB}{0, 122, 255}
\definecolor{escalator_ade}{RGB}{0, 255, 163}
\definecolor{ottoman_ade}{RGB}{255, 153, 0}
\definecolor{bottle_ade}{RGB}{0, 255, 10}
\definecolor{buffet_ade}{RGB}{255, 112, 0}
\definecolor{poster_ade}{RGB}{143, 255, 0}
\definecolor{stage_ade}{RGB}{82, 0, 255}
\definecolor{van_ade}{RGB}{163, 255, 0}
\definecolor{ship_ade}{RGB}{255, 235, 0}
\definecolor{fountain_ade}{RGB}{8, 184, 170}
\definecolor{conveyer belt_ade}{RGB}{133, 0, 255}
\definecolor{canopy_ade}{RGB}{0, 255, 92}
\definecolor{washer_ade}{RGB}{184, 0, 255}
\definecolor{plaything_ade}{RGB}{255, 0, 31}
\definecolor{swimming pool_ade}{RGB}{0, 184, 255}
\definecolor{stool_ade}{RGB}{0, 214, 255}
\definecolor{barrel_ade}{RGB}{255, 0, 112}
\definecolor{basket_ade}{RGB}{92, 255, 0}
\definecolor{waterfall_ade}{RGB}{0, 224, 255}
\definecolor{tent_ade}{RGB}{112, 224, 255}
\definecolor{bag_ade}{RGB}{70, 184, 160}
\definecolor{minibike_ade}{RGB}{163, 0, 255}
\definecolor{cradle_ade}{RGB}{153, 0, 255}
\definecolor{oven_ade}{RGB}{71, 255, 0}
\definecolor{ball_ade}{RGB}{255, 0, 163}
\definecolor{food_ade}{RGB}{255, 204, 0}
\definecolor{step_ade}{RGB}{255, 0, 143}
\definecolor{tank_ade}{RGB}{0, 255, 235}
\definecolor{trade name_ade}{RGB}{133, 255, 0}
\definecolor{microwave_ade}{RGB}{255, 0, 235}
\definecolor{pot_ade}{RGB}{245, 0, 255}
\definecolor{animal_ade}{RGB}{255, 0, 122}
\definecolor{bicycle_ade}{RGB}{255, 245, 0}
\definecolor{lake_ade}{RGB}{10, 190, 212}
\definecolor{dishwasher_ade}{RGB}{214, 255, 0}
\definecolor{screen_ade}{RGB}{0, 204, 255}
\definecolor{blanket_ade}{RGB}{20, 0, 255}
\definecolor{sculpture_ade}{RGB}{255, 255, 0}
\definecolor{hood_ade}{RGB}{0, 153, 255}
\definecolor{sconce_ade}{RGB}{0, 41, 255}
\definecolor{vase_ade}{RGB}{0, 255, 204}
\definecolor{traffic light_ade}{RGB}{41, 0, 255}
\definecolor{tray_ade}{RGB}{41, 255, 0}
\definecolor{ashcan_ade}{RGB}{173, 0, 255}
\definecolor{fan_ade}{RGB}{0, 245, 255}
\definecolor{pier_ade}{RGB}{71, 0, 255}
\definecolor{crt screen_ade}{RGB}{122, 0, 255}
\definecolor{plate_ade}{RGB}{0, 255, 184}
\definecolor{monitor_ade}{RGB}{0, 92, 255}
\definecolor{bulletin board_ade}{RGB}{184, 255, 0}
\definecolor{shower_ade}{RGB}{0, 133, 255}
\definecolor{radiator_ade}{RGB}{255, 214, 0}
\definecolor{glass_ade}{RGB}{25, 194, 194}
\definecolor{clock_ade}{RGB}{102, 255, 0}
\definecolor{flag_ade}{RGB}{92, 0, 255}

\definecolor{aeroplane_context}{RGB}{180, 120, 120}
\definecolor{bag_context}{RGB}{6, 230, 230}
\definecolor{bed_context}{RGB}{80, 50, 50}
\definecolor{bedclothes_context}{RGB}{4, 200, 3}
\definecolor{bench_context}{RGB}{120, 120, 80}
\definecolor{bicycle_context}{RGB}{140, 140, 140}
\definecolor{bird_context}{RGB}{204, 5, 255}
\definecolor{boat_context}{RGB}{230, 230, 230}
\definecolor{book_context}{RGB}{4, 250, 7}
\definecolor{bottle_context}{RGB}{224, 5, 255}
\definecolor{building_context}{RGB}{235, 255, 7}
\definecolor{bus_context}{RGB}{150, 5, 61}
\definecolor{cabinet_context}{RGB}{120, 120, 70}
\definecolor{car_context}{RGB}{8, 255, 51}
\definecolor{cat_context}{RGB}{255, 6, 82}
\definecolor{ceiling_context}{RGB}{143, 255, 140}
\definecolor{chair_context}{RGB}{204, 255, 4}
\definecolor{cloth_context}{RGB}{255, 51, 7}
\definecolor{computer_context}{RGB}{204, 70, 3}
\definecolor{cow_context}{RGB}{0, 102, 200}
\definecolor{cup_context}{RGB}{61, 230, 250}
\definecolor{curtain_context}{RGB}{255, 6, 51}
\definecolor{dog_context}{RGB}{11, 102, 255}
\definecolor{door_context}{RGB}{255, 7, 71}
\definecolor{fence_context}{RGB}{255, 9, 224}
\definecolor{floor_context}{RGB}{9, 7, 230}
\definecolor{flower_context}{RGB}{220, 220, 220}
\definecolor{food_context}{RGB}{255, 9, 92}
\definecolor{grass_context}{RGB}{112, 9, 255}
\definecolor{ground_context}{RGB}{8, 255, 214}
\definecolor{horse_context}{RGB}{7, 255, 224}
\definecolor{keyboard_context}{RGB}{255, 184, 6}
\definecolor{light_context}{RGB}{10, 255, 71}
\definecolor{motorbike_context}{RGB}{255, 41, 10}
\definecolor{mountain_context}{RGB}{7, 255, 255}
\definecolor{mouse_context}{RGB}{224, 255, 8}
\definecolor{person_context}{RGB}{102, 8, 255}
\definecolor{plate_context}{RGB}{255, 61, 6}
\definecolor{platform_context}{RGB}{255, 194, 7}
\definecolor{pottedplant_context}{RGB}{255, 122, 8}
\definecolor{road_context}{RGB}{0, 255, 20}
\definecolor{rock_context}{RGB}{255, 8, 41}
\definecolor{sheep_context}{RGB}{255, 5, 153}
\definecolor{shelves_context}{RGB}{6, 51, 255}
\definecolor{sidewalk_context}{RGB}{235, 12, 255}
\definecolor{sign_context}{RGB}{160, 150, 20}
\definecolor{sky_context}{RGB}{0, 163, 255}
\definecolor{snow_context}{RGB}{140, 140, 140}
\definecolor{sofa_context}{RGB}{250, 10, 15}
\definecolor{table_context}{RGB}{20, 255, 0}
\definecolor{track_context}{RGB}{31, 255, 0}
\definecolor{train_context}{RGB}{255, 31, 0}
\definecolor{tree_context}{RGB}{255, 224, 0}
\definecolor{truck_context}{RGB}{153, 255, 0}
\definecolor{tvmonitor_context}{RGB}{0, 0, 255}
\definecolor{wall_context}{RGB}{255, 71, 0}
\definecolor{water_context}{RGB}{0, 235, 255}
\definecolor{window_context}{RGB}{0, 173, 255}
\definecolor{wood_context}{RGB}{31, 0, 255}

\definecolor{darkgreen}{RGB}{0, 170, 40}

\definecolor{vintage_bike}{RGB}{156, 143, 189}
\definecolor{rusted_van}{RGB}{238, 147, 34}
\definecolor{fog}{RGB}{250, 112, 112}
\definecolor{green_trees}{RGB}{85, 124, 85}
\definecolor{mountains}{RGB}{25, 29, 136}
\definecolor{pastries}{RGB}{205, 141, 123}
\definecolor{wild_plate}{RGB}{104, 116, 102}
\definecolor{wooden_table}{RGB}{8, 65, 119}

\definecolor{wild_sky}{RGB}{137, 185, 173}
\definecolor{turtle}{RGB}{75, 82, 126}
\definecolor{wild_car}{RGB}{135, 35, 65}
\definecolor{wild_water}{RGB}{79, 192, 208}
\definecolor{city}{RGB}{149, 119, 119}
\definecolor{leather_bag}{RGB}{79, 158, 101}

\definecolor{dancer}{RGB}{25, 29, 136}
\definecolor{stage}{RGB}{128, 112, 112}
\definecolor{black suit}{RGB}{85, 124, 85}
\definecolor{impressionism}{RGB}{250, 112, 112}
\definecolor{theatre}{RGB}{250, 250, 0}
\definecolor{driver}{RGB}{250, 0, 0}

\definecolor{small dog}{RGB}{112, 113, 232}
\definecolor{big dog}{RGB}{198, 131, 215}
\definecolor{cabinet}{RGB}{14, 112, 112}
\definecolor{food}{RGB}{85, 124, 85}
\definecolor{white plate}{RGB}{250, 112, 112}

\definecolor{maria}{RGB}{160, 106, 100}
\definecolor{laboratory}{RGB}{112, 113, 132}
\definecolor{flask}{RGB}{198, 131, 215}

\definecolor{building_wild}{RGB}{90, 112, 112}
\definecolor{sky_wild}{RGB}{150,180, 239}
\definecolor{road}{RGB}{150, 10, 100}
\definecolor{reindeer}{RGB}{38, 147, 34}
\definecolor{snow}{RGB}{138, 147, 34}

\definecolor{luke}{RGB}{112, 113, 232}
\definecolor{leia}{RGB}{198, 131, 215}
\definecolor{ada}{RGB}{240,240, 0}
\definecolor{turing}{RGB}{250, 0, 0}

\showeditstrue

\newcommand{\oursnormal} {\texttt{CLIP-DINOiser}\xspace}
\newcommand{\ours} {\texttt{CLIP-DINOiser}\xspace}  

\newcommand{\cls} {\texttt{CLS}\xspace}

\title{\oursnormal: Teaching CLIP a few DINO tricks for open-vocabulary semantic segmentation}

\author{Monika Wysoczańska$^1$\thanks{Corresponding author: monika.wysoczanska.dokt@pw.edu.pl} \and Oriane Siméoni$^2$ \and Michaël Ramamonjisoa$^3$\thanks{Work done outside of Meta and Meta was not involved in the research discussed here.} \and Andrei Bursuc$^2$ \and $\>\>\>\>$ Tomasz Trzciński$^{1,4,5}$ $\>\>\>\>$ Patrick Pérez$^2$ 
\\
~$^1$Warsaw University of Technology, $^2$Valeo.ai,  $^3$Meta AI, $^4$Tooploox, $^5$IDEAS NCBR
  }

\begin{document}
\maketitle

\newlength{\teaserheight}
\setlength{\teaserheight}{2.4cm}
\renewcommand{\arraystretch}{0.1}
\renewcommand\cellset{\renewcommand\arraystretch{0.7}%
\setlength\extrarowheight{1pt}}
\begin{strip}
    \vspace{-65pt}
    \setlength\tabcolsep{0.0pt} 
    \centering
    \begin{tabular}{
>{\centering\arraybackslash}m{0.03\textwidth} 
>{\centering\arraybackslash}m{0.20\textwidth} 
>{\centering\arraybackslash}m{0.17\textwidth}
>{\centering\arraybackslash}m{0.20\textwidth}
>{\centering\arraybackslash}m{0.205\textwidth}
>{\centering\arraybackslash}m{0.21\textwidth}}
   \raisebox{+.1\normalbaselineskip}[0pt][0pt]{\rotatebox[origin=c]{90}{MaskCLIP}} 
   & 
 \includegraphics[height=\teaserheight, width=\linewidth]{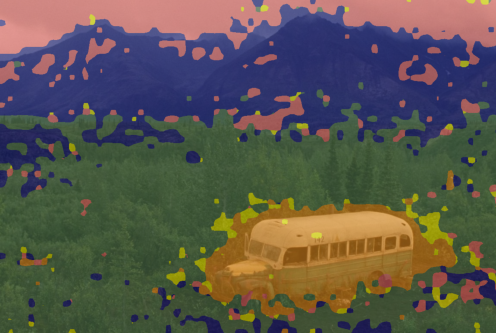}   &  \includegraphics[height=\teaserheight, width=\linewidth]{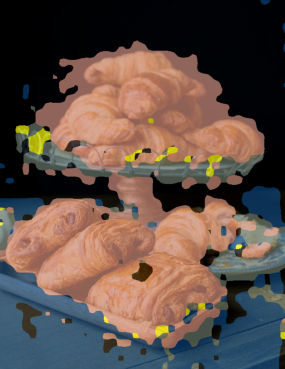} 
    &  \includegraphics[height=\teaserheight, width=\linewidth]{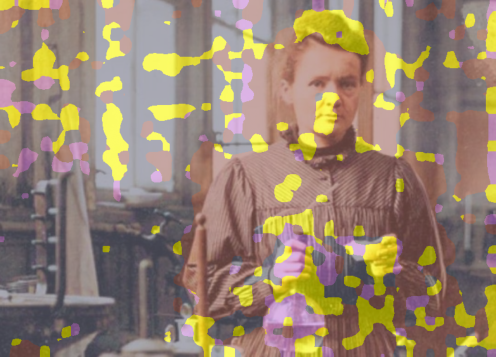} 
    & \includegraphics[height=\teaserheight, width=\linewidth,trim={0 0 4em 0}, clip]{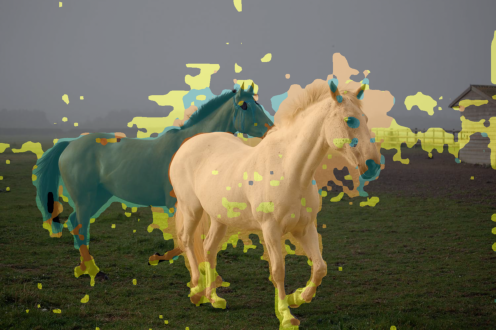}  
    & \includegraphics[height=\teaserheight, width=\linewidth]{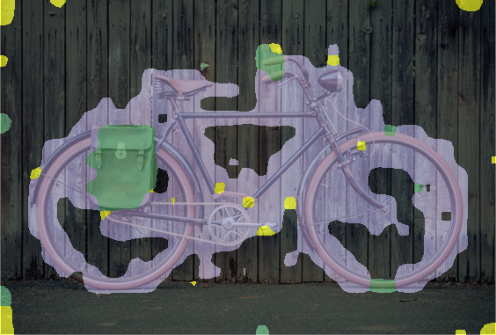} \\
   \raisebox{-0.1\normalbaselineskip}[0pt][0pt]{\rotatebox[origin=c]{90}{\bf \small{\ours}}}
   & 
 \includegraphics[height=\teaserheight, width=\linewidth]{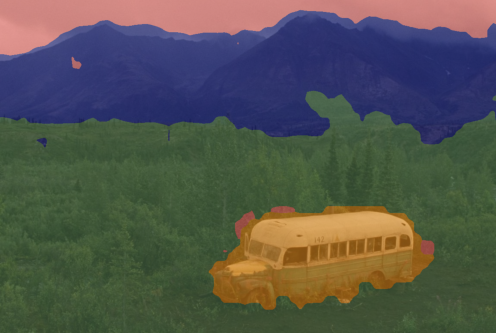}   &  \includegraphics[height=\teaserheight, width=\linewidth]{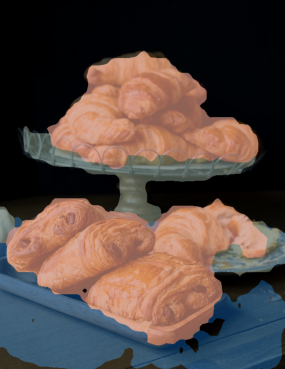} 
    &  \includegraphics[height=\teaserheight, width=\linewidth]{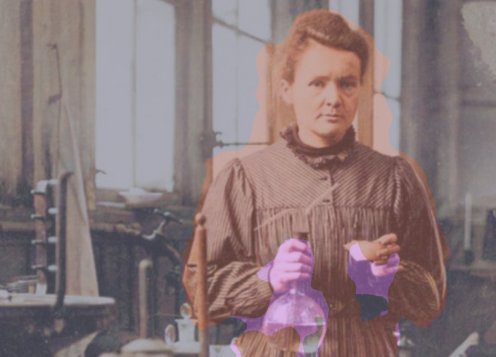} 
    & \includegraphics[height=\teaserheight, width=\linewidth,trim={0 0 4em 0}, clip]{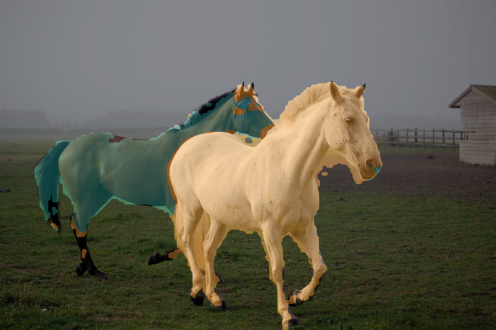}  
    & \includegraphics[height=\teaserheight, width=\linewidth]{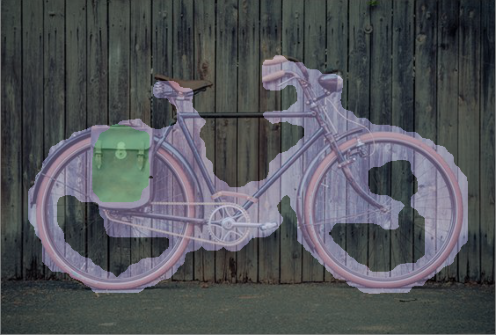} \\
    \vspace{1pt}
    & 
 {\cellcolor[gray]{.9}   
 \makecell{ \promptstyle{rusted_van}{rusted van} \\
 \promptstyle{green_trees}{green trees} \\
 \promptstyle{fog}{clouds} \promptstyle{mountains}{mountains}}}
 & 
 {\cellcolor[gray]{.9}
 \makecell{\promptstyle{pastries}{french pastries} \\
 % \promptstyle{pastries}{} \\
 \promptstyle{wooden_table}{wooden table} \\ 
 \promptstyle{wild_plate}{plate}}}
 & 
 {\cellcolor[gray]{.9}
 \makecell{
 \promptstyle{maria}{Marie Curie} \\
 \promptstyle{maria}{Sklodowska} \\
 % \promptstyle{maria}{} \\
 \promptstyle{laboratory}{laboratory}
 % \promptstyle{flask}{glass flask}}}  
 \promptstyle{flask}{flask}}}  
 & 
 {\cellcolor[gray]{.9}\makecell{\promptstyle{cell phone_coco}{white horse} \\
 \promptstyle{sky_ade}{dark horse}}}
 & 
 {\cellcolor[gray]{.9}\makecell{\promptstyle{leather_bag}{leather bag}\\
 \promptstyle{vintage_bike}{vintage bike}}} \\
 & 
 \multicolumn{5}{c}{ \large{\textcolor{Black!70!white}{l}} \cellcolor{Black!70!white} \footnotesize{\texttt{\color{white} 
 % Hallucinated
 Irrelevant prompt predicted:}} \footnotesize{\textbf{\texttt{\color{yellow} aeroplane, cat, cow, sheep, sofa, motorbike, dog}}}\large{\textcolor{Black!70!white}{l}} }
 \\
 \end{tabular}
 \vspace{-5pt}
\captionof{figure}{
\textbf{Examples of open-vocabulary semantic segmentation results} obtained with our method \ours on `in-the-wild' images vs. those of MaskCLIP~\cite{zhou2022maskclip}. Our method improves MaskCLIP features with a smart pooling strategy which does \emph{not alter the original} open-vocabulary properties. We use self-supervised DINO~\cite{caron2021dino} as a guide to \emph{teach CLIP}~\cite{openclip} to produce DINO-like localization features through two light convolutional layers.
Our method, which achieves state-of-the-art results, only requires a \emph{single forward} pass through CLIP model and our two layers. 
In addition to the correct prompts (light grey row) we list the irrelevant prompts predicted (in yellow) that we query in all images shown here.
} 
\label{fig:first}
\end{strip}

\begin{abstract} 

The popular CLIP  model displays impressive zero-shot capabilities thanks to its seamless interaction with arbitrary text prompts. However, its lack of spatial awareness makes it unsuitable for dense computer vision tasks, e.g., semantic segmentation, without an additional fine-tuning step that often uses annotations and can potentially suppress its original open-vocabulary properties. Meanwhile, self-supervised representation methods have demonstrated good localization properties without human-made annotations nor explicit supervision.
In this work, we take the best of both worlds and propose an open-vocabulary semantic segmentation method, which \emph{does not require any annotations}.
We propose to locally improve dense MaskCLIP features, which are computed with a simple modification of CLIP's last pooling layer, by integrating localization priors extracted from self-supervised features. By doing so, we greatly improve the performance of MaskCLIP and produce smooth outputs. Moreover, we show that the used self-supervised feature properties can directly be learnt from CLIP features.
Our method \ours needs only a single forward pass of CLIP and two light convolutional layers at inference,
\emph{no extra supervision nor extra memory} and reaches state-of-the-art results on challenging and fine-grained benchmarks such as COCO, Pascal Context, Cityscapes and ADE20k. 
The code to reproduce our results is available at \url{https://github.com/wysoczanska/clip_dinoiser}.
\end{abstract}      
\vspace{-15pt}
\section{Introduction}
\label{sec:intro}

Semantic segmentation is a key visual perception task for many real-world systems, e.g., self-driving cars, and industrial robots. Typically tackled in a dataset-oriented manner, best methods require a training dataset which is manually annotated for a \emph{specific and finite} set of classes.
The advent of powerful Vision-Language Models (VLM)~\cite{radford2021learning,jia2021scaling,zhai2023sigmoid} is stimulating a shift from a closed-vocabulary paradigm to an \emph{open-world} one. 
Such models are trained with a simple but scalable objective: to align pairs of image and coarse text captions that can be obtained in large amounts with limited manual supervision. 
VLMs excel at associating \emph{global} image content with arbitrary text inputs with remarkable generalization capabilities~\cite{fang2022data,mayilvahanan2023does}, but struggle to provide dense \emph{open-vocabulary features}~\cite{zhou2022maskclip,ghiasi2022openseg}.
Obtaining such an alignment between pixels and language can lead to open-vocabulary extensions for multiple other modalities, such as point clouds~\cite{jatavallabhula2023conceptfusion, peng2023openscene, chen2023clip2scene, najibi2023unsupervised}, 3D scenes~\cite{vobecky2023pop}, 3D shapes~\cite{abdelreheem2023satr}, radiance fields~\cite{kerr2023lerf}, inter-modality alignment~\cite{girdhar2023imagebind, jatavallabhula2023conceptfusion}, with multiple potential applications
for which the construction of training datasets is even more challenging and where CLIP-derived models are showing promising results.

Different strategies have been recently proposed towards improving CLIP's patch-level feature extraction abilities by modifying the original CLIP architecture for dense pooling and retraining~\cite{xu2022groupvit, cha2022tcl, clippy2022, xu2023learning, mukhoti2023pacl} or finetuning on an annotated segmentation dataset with pre-defined classes~\cite{liu2022open,zhou2022maskclip}. The former requires long training and/or large collections of annotated data, while the latter leads to an alteration of the vision-language associations of the CLIP features. An alternative line of approaches freezes the CLIP encoder and directly densifies its features with different heuristics, often with multiple forward passes~\cite{kerr2023lerf, abdelreheem2023satr, jatavallabhula2023conceptfusion, wysoczanska2023clipdiy, shin2022reco, shin2023namedmask}, but are less practical due to the extensive computational overhead.
MaskCLIP~\cite{zhou2022maskclip} arises as a computationally efficient dense CLIP extractor. It converts CLIP's global self-attention layer into a convolutional one to produce patch features with original vision-language qualities. 
If such features are local, they appear to be too noisy for high-quality segmentation mask extraction (see \cref{fig:abl-maskclip} middle column).

Meanwhile, recent self-supervised learning (SSL) approaches~\cite{caron2021dino, zhou2021ibot, mocov2, caron2020swav} produce strong visual representations displaying object localization properties, and such without requiring any manual annotation.
DINO~\cite{caron2021dino} stands out with its semantically meaningful features which have been exploited for unsupervised object discovery~\cite{simeoni2021lost, wang2022tokencut,simeoni2023found, wang2023cut}. 
DINO features prove useful also for zero-shot semantic segmentation~\cite{wysoczanska2023clipdiy, karazija2023diffusion, kerr2023lerf}, but require expensive sliding window sampling~\cite{wysoczanska2023clipdiy, kerr2023lerf} or building concept-specific prototypes and ensemble strategies~\cite{karazija2023diffusion}.

In this work, we aim for unaltered patch-level CLIP features with minimal runtime overhead. To this end, we re-examine the localization properties of MaskCLIP features and observe that it is possible to easily refine them with guidance from SSL models. 
In detail, we train a simple convolutional layer on unlabeled data to produce pooling weights to perform correlation-guided dense feature pooling from CLIP without distorting the vision-language alignment. This layer is optimized to mimic the patch correlations of DINO~\cite{caron2021dino} that indicate likely layouts of visual concepts in the images.  
Furthermore, we show that the unsupervised objectness information given by FOUND~\cite{simeoni2023found} from DINO features can be also directly learned from CLIP features again in a fully-unsupervised fashion with a single convolutional layer and helps improve the segmentation of the ill-defined `background' prompt.
With \ours, we obtain high-quality masks in \emph{a single forward pass} on CLIP (see \cref{fig:first}). 
\ours is amenable to producing dense semantic maps. 

To summarize, our contributions are: \textbf{(1)} We propose a light pooling mechanism to refine MaskCLIP features by leveraging guidance from SSL features without degrading its original open-vocabulary properties. \ours does not require any annotations, nor retraining CLIP from scratch, but only a single CLIP forward pass. \textbf{(2)} We show that CLIP \emph{already contains good localization properties} which can be exploited. We leverage simple convolutional layers to emphasize visual concept layouts from dense CLIP features. We train them without any annotation on only 1k of raw images randomly sampled in ImageNet~\cite{imagenet}. We believe that this finding could be further exploited in different contexts. \textbf{(3)} Our method achieves state-of-the-art results on complex semantic segmentation datasets such as COCO~\cite{caesar2018cocostuff}, Pascal Context~\cite{pascal-voc-2012}, Cityscapes~\cite{cordts2016cityscapes} and ADE20K~\cite{zhou2019ade20k}.

\section{Related Work}
\label{sec:related}

\parag{Zero-shot semantic segmentation.} 
This task has been typically approached by methods which aim at generalizing from \emph{seen} classes to \emph{unseen} ones~\cite{zhao2017open,xian2019semantic,bucher2019neurips,gu2020acmmm,kato2019zero,hu2020uncertainty,li2020consistent,pastore2021closer}. 
Such strategies train models with full supervision on the set of seen classes and propose different solutions to extend them to unseen ones without new images (labeled or unlabeled), e.g., by exploiting class information and relationships encapsulated in popular word embeddings~\cite{word2vec, GloVe}. While they produce fine segmentations without computational overhead, these methods require pixel-level annotations for the seen classes.

\parag{From CLIP to open-vocabulary segmentation.} 
The surge of VLMs with aligned image-language representations~\cite{radford2021learning,jia2021scaling, openclip} brought back into the spotlight the zero-shot classification task. However, the extension to zero-shot segmentation is not obvious as the CLIP architecture is not equipped to yield dense vision-language  features~\cite{zhou2022maskclip,ghiasi2022openseg}. To produce dense CLIP features, several approaches
fine-tune or train from scratch pixel-aligned CLIP-like models with additional modules, mechanisms or supervision objectives~\cite{xu2022groupvit, cha2022tcl, clippy2022, xu2023learning, mukhoti2023pacl} on datasets with annotations of varying granularity and quality: dense annotations~\cite{li2022languagedriven, liang2023open}, class-agnostic object masks~\cite{rao2021denseclip, ghiasi2022openseg, ding2021ZegFormer}, coarse captions~\cite{ghiasi2022openseg, clippy2022, zhong2022regionclip,lou2022segclip,liang2023open, xu2022groupvit,liu2022open,xu2023learning,mukhoti2023pacl, cha2022tcl} or pseudo-labels~\cite{zhou2022maskclip}. Recent works leverage image-level captions to align text to regions (obtained without supervision): PACL~\cite{mukhoti2023pacl} trains an embedder module to learn patch-to-text affinity, TCL~\cite{cha2022tcl} proposes a local constrative objective to align well-selected patches to the text and ViewCO~\cite{ren2023viewco} leverages multi-view consistency. On the downside, such models require long training on millions of images or specific types of very costly annotations.
Also, fine-tuning CLIP with a defined vocabulary is more computationally appealing~\cite{zhou2022maskclip, li2022languagedriven, liang2023open}, but alters the open-vocabulary properties of the features~\cite{jatavallabhula2023conceptfusion}. 

Most related to us is a line of works that investigate how to directly densify CLIP features~\cite{zhou2022maskclip,wysoczanska2023clipdiy,jatavallabhula2023conceptfusion,abdelreheem2023satr,kerr2023lerf} to obtain per-patch CLIP features. Such densification can be performed by aggregating features from multiple views~\cite{abdelreheem2023satr, kerr2023lerf} or from sliding windows~\cite{wysoczanska2023clipdiy,jatavallabhula2023conceptfusion} at the extra-cost of multiple forward passes.
MaskCLIP~\cite{zhou2022maskclip} drops the global pooling layer of CLIP and matches the projected features directly to text via a $1\times1$ convolution layer. By doing so they achieve dense predictions, however noisy. 

With a concept-driven perspective, some methods~\cite{shin2022reco,shin2023namedmask, karazija2023diffusion} build codebooks of visual prototypes per
concept, including negative prototypes~\cite{karazija2023diffusion}, and then perform co-segmentation~\cite{shin2022reco}. While such an approach yields good results, it is however at the cost of building expensive \emph{class-specific prototypes}, therefore diverging from open-vocabulary scenarios. 
Instead, we aim to remain \emph{open} to avoid retraining a model or building new expensive prototypes whenever a new concept is considered. To that end, we devise a dense CLIP-feature extraction method that preserves the open-vocabulary quality.

\parag{Leveraging self-supervised models \& CLIP.} Recent self-supervised ViTs~\cite{caron2021dino, zhou2021ibot, mocov2, caron2020swav, darcet2023vision} 
have demonstrated features with good localization properties~\cite{simeoni2021lost, wang2022tokencut, wang2023cut, simeoni2023found}. Such features have also been exploited in the context of open-vocabulary segmentation methods, e.g. for
pre-training for the visual backbone~\cite{clippy2022,xu2023learning, Chen_2023_ICCV}, co-segmentation~\cite{shin2022reco}, clustering patches into masks~\cite{Rewatbowornwong2023ZeroGuideSeg}, representing object prototypes~\cite{karazija2023diffusion}.
Related to us is the recent CLIP-DIY~\cite{wysoczanska2023clipdiy} which computes patch-level representations from CLIP features from different image crops with guidance from an unsupervised saliency segmenter~\cite{simeoni2023found} FOUND. While we also leverage the latter, in contrast with CLIP-DIY which runs multiple forward passes to build their dense CLIP features, our method requires only a \emph{single forward pass} of CLIP. Furthermore, our method mitigates the limits of FOUND in cluttered scenarios by integrating an uncertainty constraint. Finally, we leverage the informative patch correlation properties of DINO~\cite{caron2021dino} and show that it is possible to \emph{teach CLIP} to produce DINO-like features through light convolutional layers.
\section{Method}
\label{sec:method}

We present in this section \ours, a simple and efficient strategy to improve MaskCLIP using localization information extracted from CLIP---with a lightweight model trained to mimic some of DINO's properties. 
We first set the goal in \cref{sec:problem} and present MaskCLIP~\cite{zhou2022maskclip} in \cref{sec:denseCLIP}. 
We then introduce our strategy 
which leverages self-supervised features localization information to consolidate MaskCLIP features in \cref{sec:leveraging-self} and discuss how such localization information can directly be learnt from CLIP in \cref{sec:learning-self} (we visualize both steps in \cref{fig:method}).
We also propose a way to improve the `background' filtering in \cref{sec:background}.

\subsection{Problem statement}
\label{sec:problem}
In this work, we aim to produce open-vocabulary\footnote{We adopt the taxonomy defined in 
the recent survey~\cite{wu2024towards} and define our method as `open-vocabulary', with capabilities 
to generalize to unseen datasets.} semantic segmentation of an image. 
We consider an image $X \in \mathbb{R}^{H \times W \times 3}$ which we split into a sequence of $N$ patches of dimensions $P \times P \times 3$ with $P\times P$ the patch size and $N = \lceil \frac{H}{P}\rceil \cdot \lceil\frac{W}{P}\rceil$. A class token, noted \texttt{CLS}, is added to the input sequence and we feed the $N+1$ patches to a ViT~\cite{dosovitskiy2020image} model. We aim at producing dense visual features $F \in \mathbb{R}^{N \times d}$, with $d$ the feature dimension, that can later be matched to \emph{any} set of text inputs embedded in the same space. In particular, 
the goal is to produce a segmentation map per textual query.

\subsection{Preliminaries on MaskCLIP} 
\label{sec:denseCLIP}

\parag{Extracting dense open-vocabulary features.}
The popular CLIP~\cite{openclip} model pre-trained on image/caption pairs produces good \emph{global} image features, but was not trained to generate high-quality 2D feature maps.
In order to extract such dense feature maps relevant to semantic segmentation, Zhou et al.~\cite{zhou2022maskclip} revisit the global attention pooling layer of the last attention layer of the model.
% Indeed, 
The authors discard the \emph{query} and \emph{key} embeddings of the 
layer and transform both the \emph{value} projection and the last linear layer into a conv $1\times1$ layer. With this new model, named MaskCLIP and denoted $\phi(\cdot)$, we extract 
$d$-dimensional features $\phi^L(X) \in \mathbb{R}^{N \times d}$ from the last layer $L$ which retains most of the open-vocabulary properties of CLIP~\cite{zhou2022maskclip}.

\parag{Semantic segmentation given textual queries.}
We also extract CLIP textual features $\phi_T(t_j)$ for each 
text query $t_j \in \mathcal{T}$ with $j \in \{1,\ldots,|\mathcal{T}|\}$. Segmentation maps are then generated by computing the cosine similarity between each of the visual patch features and of the textual prompts, after L2-normalization. The most similar prompt is assigned to each patch. Note that a query `background' can be added in order to obtain \emph{negative} patches.
Using MaskCLIP allows us to produce dense segmentation maps with a single forward pass of the classic CLIP model, but its outputs are noisy, as visible in \cref{fig:abl-maskclip} (middle column). 

\vspace{-5pt}
\subsection{DINOising open-vocabulary features}
\label{sec:leveraging-self}
In this work, we aim to improve MaskCLIP's open-vocabulary features described above. To do so, we propose 
to leverage the known good localization properties of self-supervised features~\cite{caron2021dino, oquab2023dinov2, simeoni2021lost, wang2022tokencut, simeoni2023found, simeoni2023survey} .

\parag{Extracting self-supervised correlation information.}
Recent works~\cite{simeoni2021lost, wang2022tokencut} have shown that the patch correlation information of the embeddings from the last attention layer of the self-supervised model, DINO~\cite{caron2021dino} can help highlight objects in images. We use here the \emph{value} embeddings which we observe have finer correlation than those of key and query (more discussion in  \cref{sec:self-abl}).
We extract such self-supervised features $\xi(X)\in\mathbb{R}^{N\times d_\xi}$ and discard the \cls token.
We then compute the per-patch cosine-similarity and produce the affinity map $\dinocorr \in [-1,1]^{N \times N}$.
We compare in \cref{fig:corr-comparison} the patch-similarities obtained for a patch \emph{seed} with MaskCLIP and DINO features and observe that the self-supervised features are more densely and accurately correlated than those of CLIP.

\begin{figure}[!ht]
\centering
    \includegraphics[width=.9\linewidth, trim={8em, 1.2em, 8em, 0}, clip] {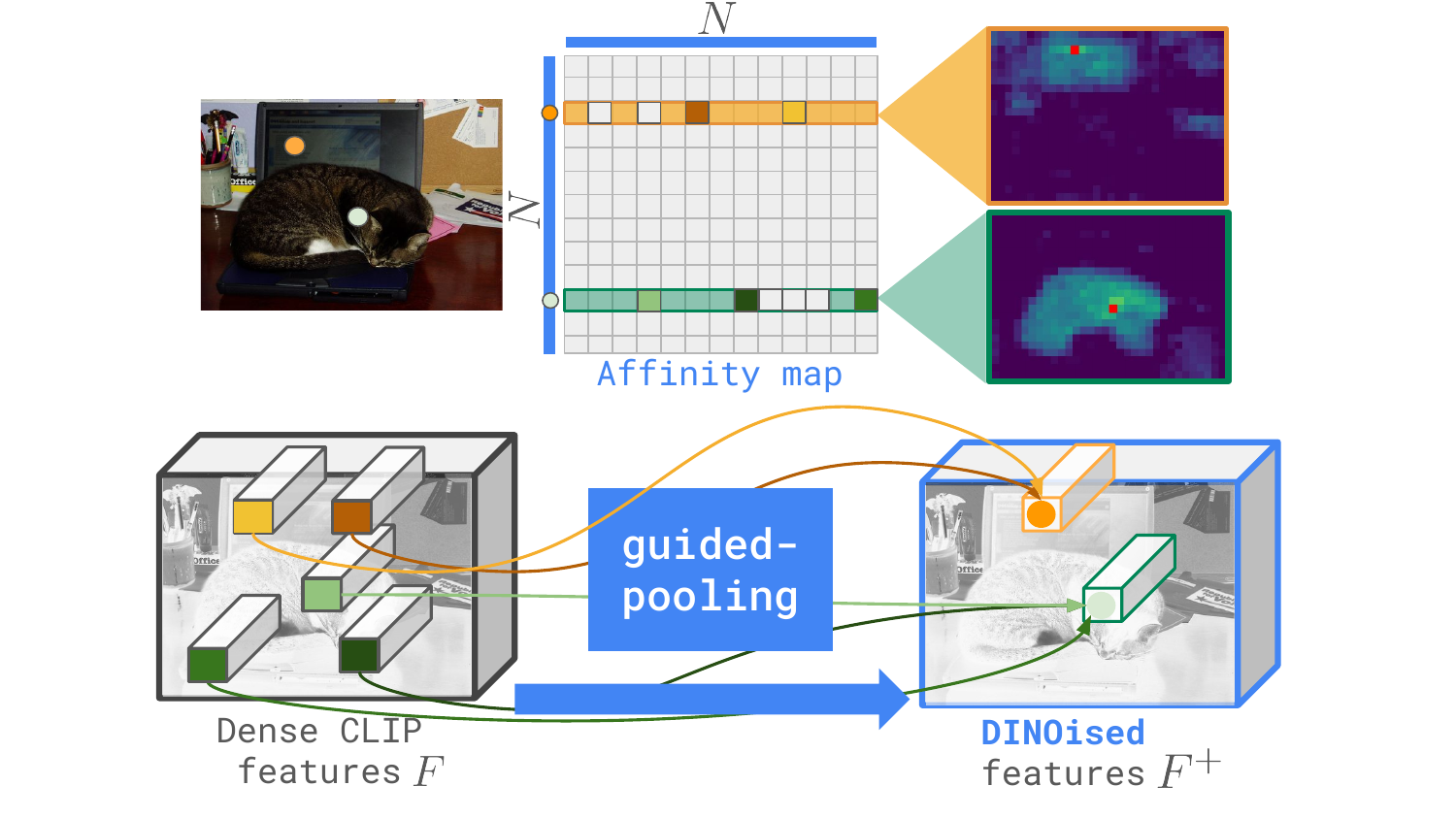}
\vspace{-5pt}
\caption{\textbf{Guided pooling} 
strategy defined in Eq.~\eqref{eq:refinement}.
The $N\times N$ affinity matrix is computed from patch features and is used to refine MaskCLIP features (bottom left).
}
\label{fig:corr-zoom}
\vspace{-10pt}
\end{figure}

\parag{Strengthening features with guided pooling.}
In order to locally consolidate MaskCLIP features $\clipfeat$, now noted $F$, we propose to perform a \emph{concept-aware} linear combination of the features per patch with guidance from the patch affinity $\dinocorr$. 
The feature combination strategy can be seen as a form of voting mechanism that enforces similar patches to have similar CLIP features (and prediction) while attenuating noisy features. 
Specifically, we compute the new features $\clippatch{}^{+}\in \mathbb{R}^{N \times d}$ as an average of MaskCLIP features $F$ weighted by $\dinocorr$, presented in \cref{fig:corr-zoom}. 
We zero-out $\dinocorr$ correlations below a threshold $\gamma$, following~\cite{simeoni2021lost, wang2022tokencut}, and compute
the new features for patch $p \in \{1,\ldots,N\}$:
\begin{equation}
\clippatch{p}^{+} = \frac{1}{\sum_{q=1}^{N} A^\xi_{p,q}}\sum_{q=1}^{N} A^\xi_{p,q} \cdot F_p.
\label{eq:refinement}
\end{equation}

\newlength{\mainpoolingheight}
\setlength{\mainpoolingheight}{1.7cm}
\renewcommand{\arraystretch}{0.2}

\setlength{\mainpoolingheight}{1.7cm}
\renewcommand{\arraystretch}{0.2}
\begin{figure}[!ht]
\centering
    \centering
    \setlength\tabcolsep{0.5pt}
    \begin{tabular}{cccc}
    & \small{GT} & \small{using $F$} & \small{using $F^{+}$} \\
    \raisebox{+2.1\normalbaselineskip}[0pt][0pt]{\rotatebox[origin=c]{90}{\small{Context}}} & \includegraphics[width=0.3\linewidth]{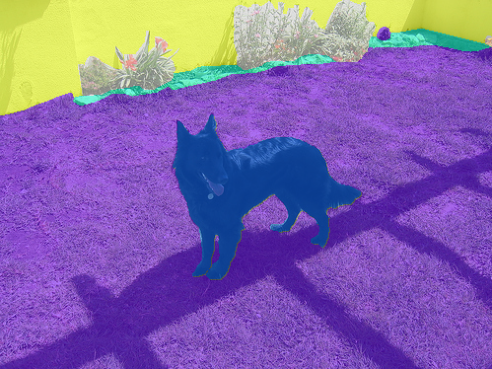} & 
    \includegraphics[width=0.3\linewidth]{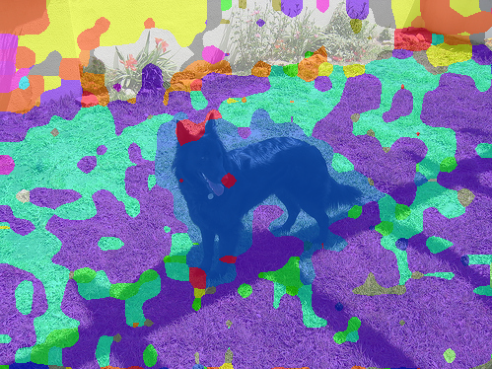} & 
    \includegraphics[width=0.3\linewidth]{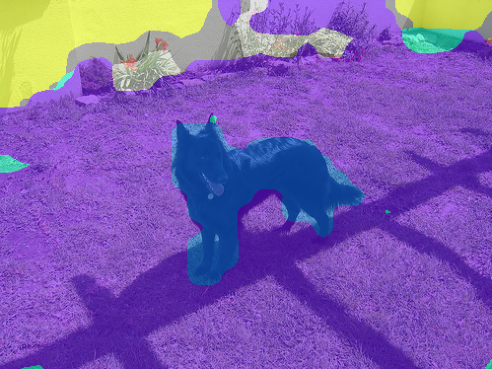}  \\ 
    
    \raisebox{+1.6\normalbaselineskip}[0pt][0pt]{\rotatebox[origin=c]{90}{\small{ADE20k}}} & \includegraphics[width=0.3\linewidth]{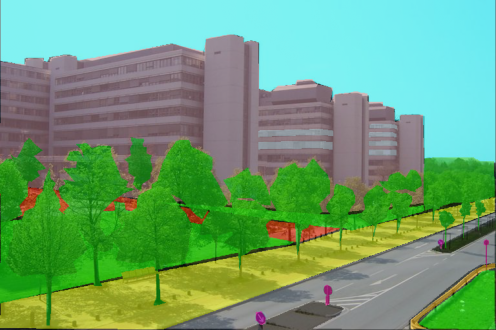} & 
    \includegraphics[width=0.3\linewidth]{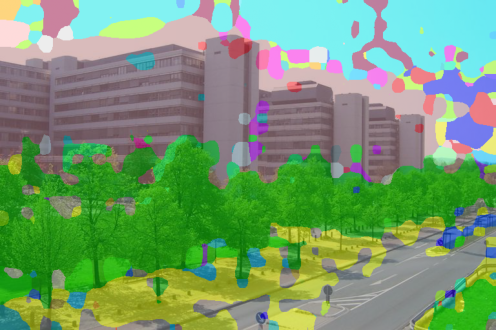} & 
    \includegraphics[width=0.3\linewidth]{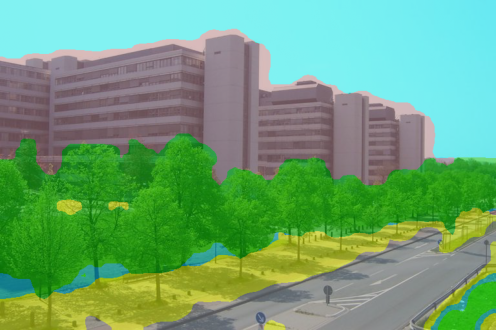}  \\
    \end{tabular}
\vspace{-5pt}
\caption{\textbf{Impact of the pooling.} 
We compare our 
results with $F^{+}$ (right) versus those obtained with 
MaskCLIP features (middle). 
}
\vspace{-8pt}
\label{fig:abl-maskclip}
\end{figure}

We then produce the segmentation maps $S \in [-1,1]^{N \times |\mathcal{T}|}$, by comparing the new features $\clippatch{}^{+}$ to each textual queries in $\mathcal{T}$. 
As shown in \cref{fig:abl-maskclip}, when using such consolidated features, we obtain more 
accurate outputs and the high-frequency predictions observed in MaskCLIP are smoothed out, showing the benefit of the pooling.

\subsection{Teaching CLIP a first DINO trick: object correlations}
\label{sec:learning-self}

We have shown in the previous section that self-supervised correlation information can successfully be used to improve the dense quality of open-vocabulary features. If the difficulty of densifying CLIP is well-known, we show here that CLIP features already contain \emph{good localization information} which can be extracted with a light model. 
We indeed predict
DINO correlations $A^\xi$ from CLIP with a single convolutional layer.

\newlength{\corrwidth}
\setlength{\corrwidth}{0.33\linewidth}
\newlength{\corrheight}
\setlength{\corrheight}{1.6cm}

\setlength{\mainpoolingheight}{1.7cm}
\renewcommand{\arraystretch}{0.2}
\begin{figure}[!ht]
\centering

\vspace{-5pt}

\setlength\tabcolsep{1pt} 
\centering
\resizebox{\linewidth}{!}{
\begin{tabular}{
cccc
}
   \small{image} &  \small{MaskCLIP corr.}  & \small{DINO  $\dinocorr$} & \small{ours $\clipcorr$} \\
   \includegraphics[width=0.24\linewidth, height=\corrheight]{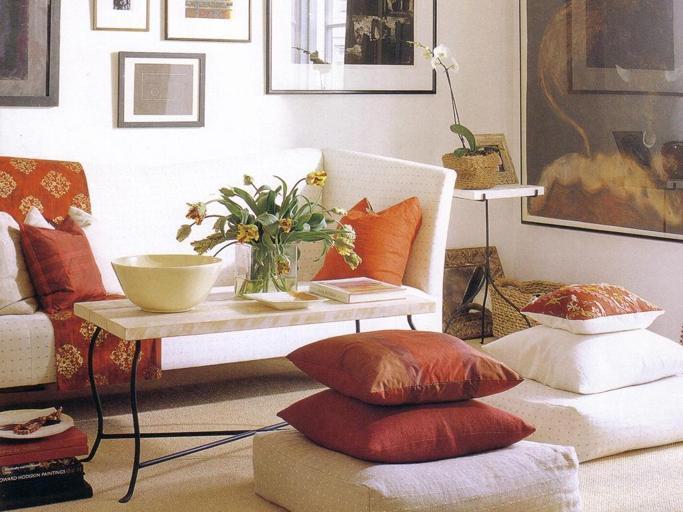}
   & 
   \includegraphics[width=0.24\linewidth, height=\corrheight]{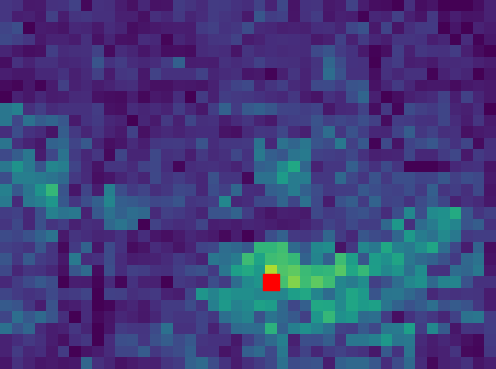}
   &
   \includegraphics[width=0.24\linewidth, height=\corrheight]{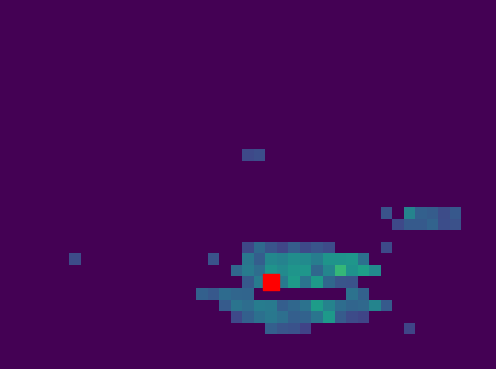} 
    &  \includegraphics[width=0.24\linewidth, height=\corrheight]{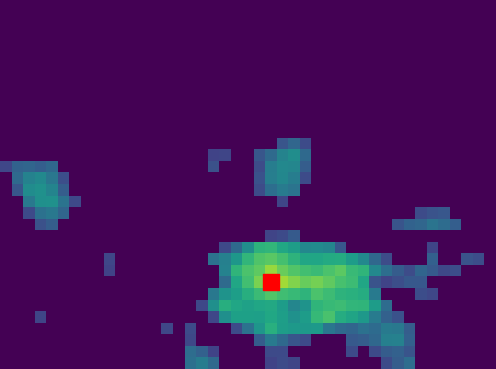} 
\end{tabular}
}
\vspace{-8pt}
\caption{
\textbf{Comparison of the affinity maps} between a \textcolor{red}{\emph{seed}} (on a `pillow') 
and the other patch features when using
features of MaskCLIP, DINO and ours after training.
}
\label{fig:corr-comparison} 
\vspace{-5pt}
\end{figure}

\begin{figure*}[ht!]
    \centering
    \includegraphics[width=0.9\linewidth, trim={0em, 9.2em, 4.5em, 0em}, clip]{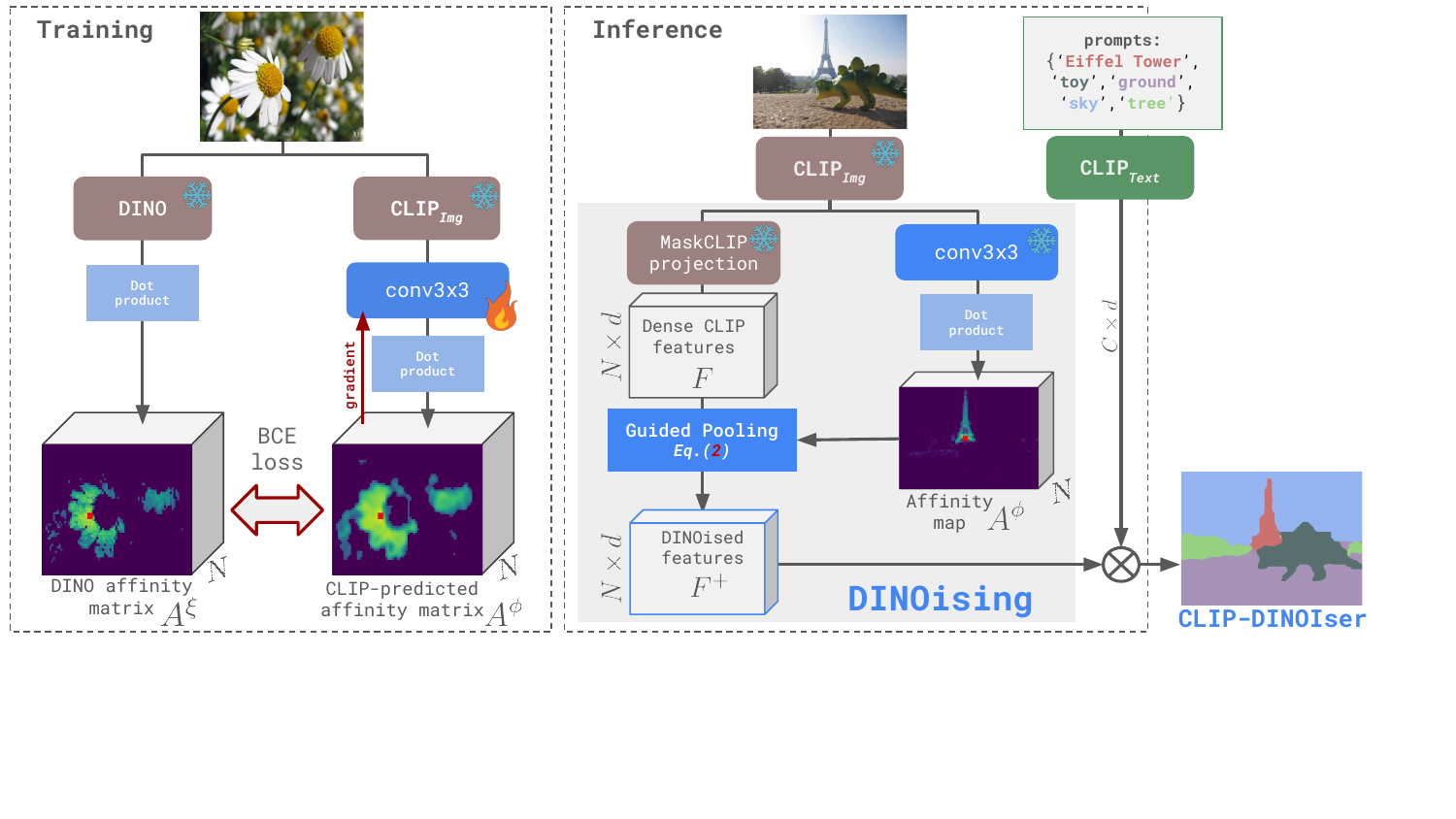}
    \vspace{-8pt}
    \caption{\textbf{Overview of \ours} which leverages the quality of self-supervised features to improve the notoriously noisy MaskCLIP feature maps. 
    We use DINO as a teacher which `teaches' CLIP how to extract localization information. We train (left) a conv$3\times3$ layer to reproduce the patch correlations obtained with DINO.
    At inference (right), an input image is forwarded through the frozen CLIP image backbone and MaskCLIP projection. The produced features are then improved with our \emph{pooling} strategy which is guided by correlations predicted with the trained convolutional layer applied on CLIP. With this light `DINOising' process, we obtain `DINOised' features which are matched against the prompts features to produce \ours outputs.
    }
    \vspace{-10pt}
    \label{fig:method}
\end{figure*}

In order to predict the DINO affinity map $\dinocorr$ from CLIP features, we train a \emph{single $3 \times 3$ convolutional layer} $g(\cdot):\mathbb{R}^d \rightarrow \mathbb{R}^{d_g}$ which projects intermediate features $\clipinterfeat$--extracted from layer $l$--into a smaller space of dimension $d_g < d$. We enforce the patch correlations of the generated features $\clipcorr \in [-1,1]^{N \times N}$:
\begin{equation}
\clipcorr = \frac{g(\clipinterfeat)}{\|g(\clipinterfeat)\|} \otimes \left(\frac{g(\clipinterfeat)}{\|g(\clipinterfeat)\|}\right)^\top,
\label{eq:correlation2}
\end{equation}
with $\otimes$ denoting the outer product, 
to be close to the binarized correlations $D = A^\xi > \gamma $ (we use here the same $\gamma$ as defined above), using the binary cross-entropy loss $\mathcal{L}^c$:
\begin{equation}
   \mathcal{L}^c = \sum_{p=1}^N \left[ D_{p} \log A^\phi_{p} + (1 - D_{p}) \log ( 1 - A^\phi_{p}) \right].
   \label{eq:loss_dino_corr}
\end{equation}

We present our layer training in \cref{fig:method} (left part) and observe the quality of CLIP-predicted affinity matrix $A^\phi$.
%}.
We also show in \cref{fig:corr-comparison} another example of obtained $A^\phi$ and observe their similarity to DINO-based correlations.
We use the CLIP-produced correlations $A^\phi$ to replace $A^\xi$ in \cref{eq:refinement} to weight the pooling and observe a similar boost over MaskCLIP, 
thus showing that good patch correlations can indeed be extracted directly from CLIP. We can now discard DINO and we name \ours the guided-pooling strategy which uses CLIP-based correlation.
As shown in \cref{fig:method} (\emph{inference} step), our method runs with a single forward pass of CLIP model and a small extra layer.

\subsection{Teaching CLIP a second DINO trick: background filtering}
\label{sec:background}

Moreover, as discussed earlier, a `background' query may be added to the set of textual queries $\mathcal{T}$ in order to help filter out patches falling in the \emph{background} and not corresponding to any objects. We do not assume here any prior knowledge about classes of interest and focus rather on the foreground/background paradigm \cite{simeoni2023found}.
We argue that relying solely on the textual prompt `background' to catch all non-salient patches is underperforming and, similarly to~\cite{wysoczanska2023clipdiy}, we propose to use a very light-weight \emph{unsupervised} foreground/background segmentation method, namely FOUND~\cite{simeoni2023found} which also relies on DINO self-supervised features. We run FOUND on the entire image and extract a prediction mask $M \in \{0,1\}^{N}$ in which a patch is assigned the value $1$ if falling into the foreground and $0$ otherwise. 
We also observe that saliencies produced by FOUND can be too restrictive and discard objects which are partially visible or in a clutter. In order to mitigate this behaviour, we propose to relax the background selection by integrating an additional uncertainty constraint.
To this end, we fuse the background information from both modalities by assigning
the `background' prompt to patches $p$ which are both \emph{uncertain}, e.g. have low confidence score $\sigma(S)_p < \delta$, with $\sigma(\cdot)$ the softmax operation, \emph{and} which fall in the background in $M$.

\begin{figure}[h!]  
    \vspace{-3pt}
    \renewcommand{\arraystretch}{0.15}
    \setlength\tabcolsep{0.9pt} 
    \centering
    \resizebox{\linewidth}{!}{
    \begin{tabular}{cc}
          \small{FOUND~\cite{simeoni2023found}} & \small{ \oursnormal} \\
         \includegraphics[height=1.7cm] {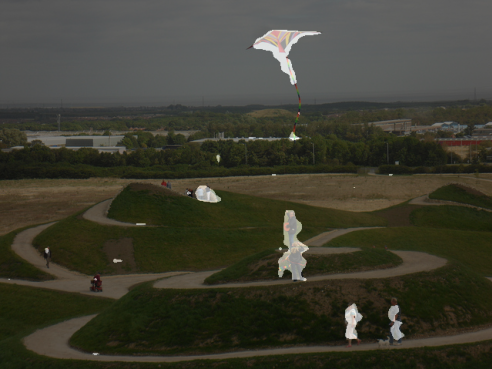}\includegraphics[height=1.7cm]{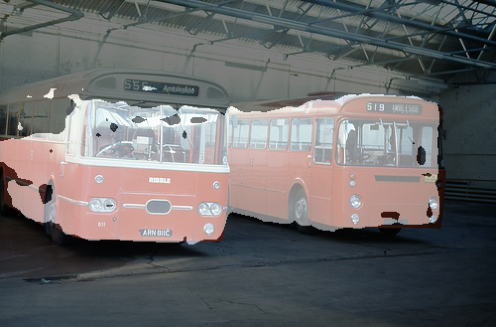} & \includegraphics[height=1.7cm]{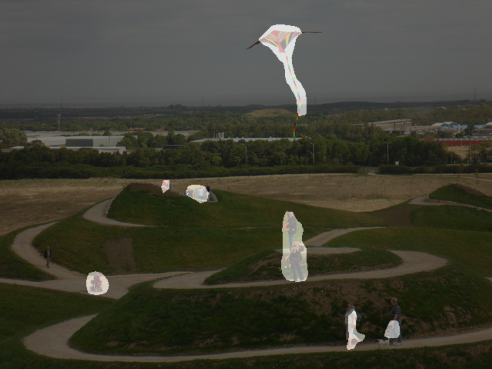}\includegraphics[height=1.7cm]{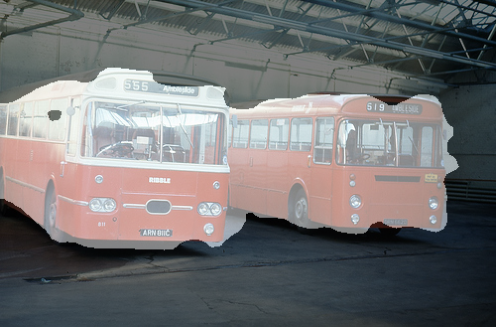} \\
    \end{tabular}
    }
    \vspace{-9pt}
  \caption{\textbf{Comparison of \emph{objectness} mask} generated by FOUND~\cite{simeoni2023found} (left) and with our layer using CLIP features (right).}
  \vspace{-5pt}
  \label{fig:found-pred}
\end{figure}

\begin{figure}[h!]
    \centering
    \includegraphics[width=0.98\linewidth, trim={2.9em, 14em, 37em, 1.5em}, clip]{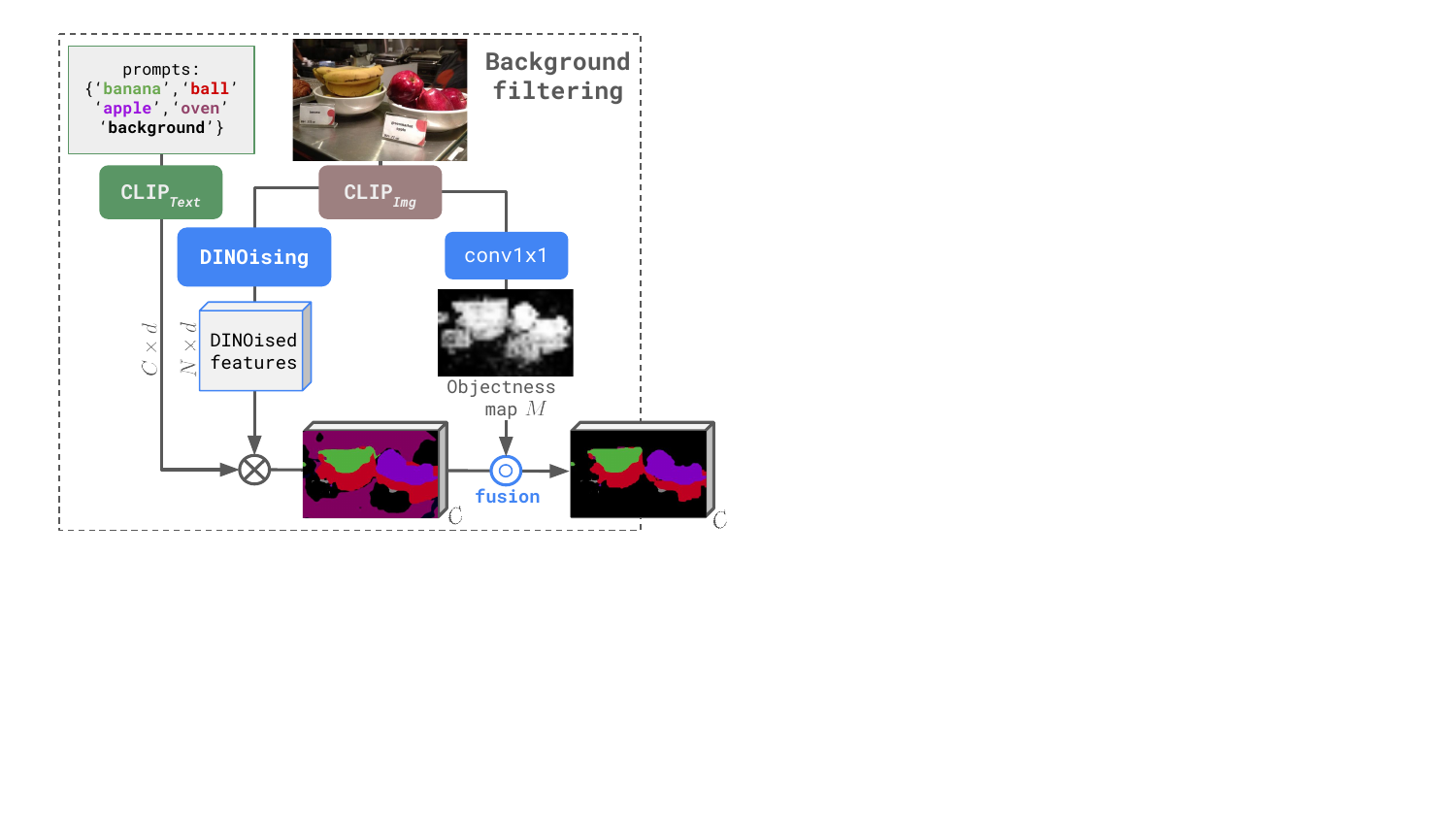}   
    \caption{\textbf{Overview of our \emph{background filtering}} applied when a `background' prompt is provided to help reduce hallucinations.} 
    \vspace{-15pt}
    \label{fig:back-filter}
\end{figure}

\parag{Learning FOUND objectness.} Moreover, we are also able to 
learn the predictions of FOUND~\cite{simeoni2023found} directly from CLIP features. To do so, we train a \emph{single $1 \times 1$} convolutional layer $h(\cdot):\mathbb{R}^d \rightarrow \mathbb{R}$ which predicts from the features $\clipinterfeat$ an objectness map $M^\phi = h(\clipinterfeat) \in \mathbb{R}^{N}$. We train the model to predict the FOUND binary mask $M$ with the binary cross-entropy loss $\mathcal{L}^m$:
\begin{equation*}
   \mathcal{L}^m = \sum_{p=1}^N \left[ M_{p} \log(M^\phi_p) + (1 - M_{p}) \log(1 - M^\phi_p) \right].
   % Removed
   % \label{eq:loss_dino_corr2}
\end{equation*}
We show examples of predicted CLIP-based objectness in \cref{fig:found-pred} and observe their very high similarity to those produced with DINO. Moreover, we can now replace $M$ defined above with the binarized CLIP-based scores $\zeta(M^\phi) > 0.5$, with $\zeta(\cdot)$ the sigmoid operation, and observe a minimal drop in performances. We provide an example of the \emph{background filtering} with trained objectness in \cref{fig:back-filter}.

\section{Experiments}
\label{sec:experimental}

We detail in \cref{sec:setup} the experimental setup used in our evaluation. We produce state-of-the-art results on the task of open-vocabulary semantic segmentation in \cref{sec:zero-shot} and ablation studies in \cref{sec:ablation}. 

\subsection{Experimental setup}
\label{sec:setup}

\parag{Technical details.} We use in all experiments a \emph{frozen} CLIP ViT-B/16 pre-trained following OpenCLIP~\cite{openclip}.
Our method \ours uses two convolutional layers to extract DINO-like information from CLIP layer $l=10$ (the 3rd before the last which was shown to provide the best results \cite{Walmer_2023_CVPR}). The first layer $g(\cdot)$ has a kernel $3\times3$ and output dimension 
$d_g = 256$ and $h(\cdot)$ a kernel $1\times1$ with $d_h = 1$. The first is trained to match the correlation information extracted from the \emph{value} embeddings of the last layer of a ViT-B/16 model trained following DINO~\cite{caron2021dino}.
The second layer is trained to replicate the unsupervised object localization predictions of FOUND~\cite{simeoni2023found}--which also uses DINO model. We train both layers with a binary cross-entropy loss on \emph{only 1k raw images} randomly sampled from ImageNet~\cite{imagenet} dataset \emph{without any annotation}.
We report average scores over 3 runs with different sampling seeds and provide standard deviations in appendix (\cref{sec:sup_exp}).
We follow \cite{wang2022tokencut} and binarize the correlations with $\gamma=0.2$. In the background filtering step, we use a high confidence score, i.e., $\delta=0.99$.
We train our model for 6k iterations with a batch size of 16 images using Adam optimizer~\cite{kingma2015}, which takes approximately 3 hours on a single NVIDIA RTX A5000 GPU. We decrease the learning rate for both heads by a factor of 0.1 after 5k iterations. We apply data augmentations during training (random scale and cropping, flipping and photometric distortions).  

\vspace{-5pt}
\parag{Datasets and metric.}
We evaluate our method on eight benchmarks typically used for zero-shot semantic segmentation~\cite{cha2022tcl}. Following~\cite{cha2022tcl}, we split them into two groups. The first consists in datasets with a `background' query: PASCAL VOC~\cite{pascal-voc-2012} (noted `VOC'), PASCAL Context~\cite{mottaghi_cvpr14} (noted `Context'), and COCO Object~\cite{caesar2018cocostuff} (noted `Object') and the second without: PASCAL VOC20~\cite{pascal-voc-2012} (noted `VOC20'), PASCAL Context59~\cite{mottaghi_cvpr14} (noted `C59'), COCO-Stuff~\cite{caesar2018cocostuff} (noted `Stuff'),
Cityscapes~\cite{cordts2016cityscapes} (noted `City'), and ADE20K~\cite{zhou2019ade20k} (noted `ADE').  We evaluate results with the standard mIoU metric. We also follow the evaluation protocol of~\cite{cha2022tcl}, use the implementations provided by MMSegmentation~\cite{mmseg2020}, employ a sliding window strategy, resize the input image to have a shorter side of 448.
We also do not perform text expansions of the class names and 
use only the standard ImageNet prompts following \cite{openclip, xu2022groupvit, zhou2022maskclip}. 

\vspace{-3pt}
\parag{Baselines.} We compare our method against state-of-the-art methods on open-vocabulary zero-shot semantic segmentation. For a fair comparison between methods, we report results without any post-processing step. In our evaluations, we follow the taxonomy presented in~\cite{wu2024towards} and compare our model with the methods relying on language-image pretraining, also called open-vocabulary. We split the compared baselines into four categories: (1) \emph{dataset specific} which employ pseudo-labeling and supervised training of a segmentation model on target dataset: NamedMask~\cite{shin2023namedmask}, MaskCLIP+~\cite{zhou2022maskclip}); (2) \emph{construct prototypes}: ReCO~\cite{shin2022reco}, OVDiff~\cite{karazija2023diffusion}; (3) \emph{train with text supervision} including GroupViT~\cite{xu2022groupvit}, ZeroSeg~\cite{Rewatbowornwong2023ZeroGuideSeg}, SegCLIP~\cite{lou2022segclip}, TCL~\cite{cha2022tcl}, CLIPpy~\cite{clippy2022}, OVSegmentor~\cite{xu2023learning}, which all require access to additional datasets of millions of image/caption pairs (we note in the table the exact datasets used for the training); and finally \emph{use frozen CLIP} i.e. CLIP-DIY~\cite{wysoczanska2023clipdiy} and MaskCLIP~\cite{zhou2022maskclip}, which use pre-trained CLIP. Our method falls into the last category as we do not modify CLIP, and do not need access to additional caption annotations as we use only 1k unannotated images.

\newcommand{\resultsrownobg}[9]{#1 & #2 & #3 & #4 & #5 & #6 & #7 & #8 & #9 & \more}
\newcommand{\more}[4]{#1 & #2 & #3 & #4 }

\begin{table*}[ht]
\centering
\small
\renewcommand{\arraystretch}{1.0}
\setlength{\tabcolsep}{1.2pt}
% \resizebox{\linewidth}{!}{
\begin{tabular}
{l|cccc|ccccc|ccc}
\toprule

& \textbf{Concept} &
{\textbf{Frozen}} &
{\textbf{Extra}} &
{\textbf{Backbone}} &
\multicolumn{5}{c}{\bf No background prompt} &
\multicolumn{3}{c}{\bf W/ bkg prompt} 
\\
\resultsrownobg
{\textbf{Methods}}
{\textbf{spec.}}
{\textbf{backbone}}
{\textbf{data}}
{\textbf{at inference}}
{VOC20}
{C59}
{Stuff}
{City}
{ADE}
{Context}
{Object}
{VOC} \\

\midrule
\rowcolor{Black!10!white} \multicolumn{13}{l}{\textcolor{Black!10!white}{aaaaaaa} \textbf{Dataset specific}} \\
\resultsrownobg
{MaskCLIP+~\cite{zhou2022maskclip}}
{\checkmark}
{\xmark}
{Target dataset}% {Pseudo labels}
{DeepLabv2}
{-} % VOC20
{-} % Context59
{18.0} % Stuff
{-} % City
{-} %ADE 
{\underline{31.1}} % Context
{-} % Object 
{-} % VOC
% 31.1 18.0
\\ 

\resultsrownobg{NamedMask~\cite{shin2023namedmask}}
{\checkmark}
{\xmark}
{IN(1.2M)+Target}
{DeepLabv3+}
{-}
{-}
{-}
{-}
{-}
{-} % Context
{27.7} % Object
{59.2} % VOC
\\

\midrule
\rowcolor{Black!10!white} \multicolumn{13}{l}{\textcolor{Black!10!white}{aaaaaaa} \textbf{Build prototypes per visual concept}} \\

\resultsrownobg
{ReCo~\cite{shin2022reco}}
{\checkmark}
{\checkmark}
{IN(1.2M)}
{CLIP}
{57.8} % VOC20
{22.3} % Context59
{14.8} % Stuff
{21.1} % City
{11.2} 
{19.9} % Context
{15.7} % Object 
{25.1} % VOC\\ 
\\

\resultsrownobg
{OVDiff~\cite{karazija2023diffusion}}
{\checkmark}
{\checkmark}
{\xmark}
{CLIP+DINO+SD}
{\bf 81.7} % VOC20
{\underline{33.7}} % Context59
{-} % Stuff
{-} % City
{\underline{14.9}} 
{30.1} % Context
{\bf 34.8} % Object
{\bf 67.1} % VOC
\\ %ADE 

\midrule
\rowcolor{Black!10!white} \multicolumn{13}{l}{\textcolor{Black!10!white}{aaaaaaa} \textbf{Text/image alignment training with captions}} \\

\resultsrownobg
{GroupViT~\cite{xu2022groupvit}}
{\xmark}
{\xmark}
{CC12M~\cite{changpinyo2021conceptual}+RedCaps~\cite{desai2021redcaps}}
{CLIP}
{79.7} % VOC20
{23.4} % Context59
{15.3} % Stuff
{11.1} % City
{9.2}
{18.7} % Context
{27.5} % Object 
{50.4} % VOC
\\ %ADE

\resultsrownobg{ZeroSeg~\cite{Chen_2023_ICCV}}
{\xmark}
{\xmark}
{IN(1.2M)+CC12M~\cite{changpinyo2021conceptual}}
{CLIP}
{-}
{-}
{-}
{-}
{-}
{21.8} % Context
{22.1} % Object 
{42.9} % VOC
\\

\resultsrownobg
{SegCLIP~\cite{lou2022segclip}}
{\xmark}
{\xmark}
{CC3M~\cite{cc2018}+COCO(400k)}
{CLIP}
{-} % VOC20
{-} % Context59
{-} % Stuff
{11.0} % City
{8.7} 
{24.7} % Context
{26.5} % Object 
{52.6} % VOC
\\ %ADE 

\resultsrownobg
{TCL~\cite{cha2022tcl}}
{\xmark}
{\xmark}
{CC12M~\cite{changpinyo2021conceptual}+CC3M~\cite{cc2018}}
{CLIP}
{77.5} % VOC20
{30.3} % Context59
{\underline{19.6}} % Stuff
{23.1} % City
{\underline{14.9}} 
{24.3} % Context
{30.4} % Object 
{51.2} % VOC
\\ %ADE 

\resultsrownobg
{CLIPpy~\cite{clippy2022}}
{\xmark}
{\xmark}
{HQITP-134M~\cite{clippy2022}}
{CLIP}
{-} % VOC20
{-} % Context59
{-} % Stuff
{-} % City
{13.5} 
{-} % Context
{32.0} % Object 
{52.2} % VOC
\\ %ADE 

\resultsrownobg
{OVSegmentor~\cite{xu2023learning}}
{\xmark}
{\xmark}
{CC4M~\cite{xu2023learning}}
{CLIP}
{-} % VOC20
{-} % Context59
{-} % Stuff
{-} % City
{5.6} 
{20.4} % Context
{25.1} % Object 
{53.8} % VOC
\\ %ADE 

\midrule
\midrule

\rowcolor{Black!10!white} \multicolumn{13}{l}{\textcolor{Black!10!white}{aaaaaaa} \textbf{Frozen CLIP}} \\

\resultsrownobg
{CLIP-DIY~\cite{wysoczanska2023clipdiy}$^{*}$}
{\xmark}
{\checkmark}
{\xmark}
{CLIP+DINO}
{79.7} % VOC20
{19.8} % Context59
{13.3} % Stuff
{11.6} % City
{9.9} 
{19.7} % Context
{31.0} % Object 
{59.9} % VOC
\\ %ADE 

\resultsrownobg
{MaskCLIP~\cite{zhou2022maskclip}~\cite{cha2022tcl}}
{\xmark}
{\checkmark}
{\xmark}
{CLIP}
{53.7} % VOC20
{23.3} % Context59
{14.7} % Stuff
{21.6} % City
{10.8} 
{21.1} % Context
{15.5} % Object 
{29.3} % VOC
\\ %ADE 

\resultsrownobg
{MaskCLIP$^{*}$} %(OpenCLIP Laion 2B)}
{\xmark}
{\checkmark}
{\xmark}
{CLIP}
{61.8} % VOC20
{25.6} % Context59
{17.6} % Stuff
{\underline{25.0}} % City
{14.3} 
{22.9} % Context 
{16.4} % Object
{32.9} % VOC
\\ %ADE 

\resultsrownobg
{MaskCLIP$^{*}$ $\dag$} %(OpenCLIP Laion 2B)}
{\xmark}
{\checkmark}
{\xmark}
{CLIP}
{71.9} % VOC20
{27.4} % Context59
{18.6} % Stuff
{23.0} % City
{\underline{14.9}} 
{24.0} % Context
{21.6} % Object
{41.3} % VOC
\\ %ADE

\resultsrownobg
{\rowcolor{ForestGreen!15!white} \oursnormal}
{\xmark}
{\checkmark}
{IN (random 1k im.)}
% {\xmark}
{CLIP}
{\underline{80.9}} % VOC20
{\bf35.9} % Context59
{\bf24.6} % Stuff
{\bf31.7} % City
{\bf20.0} 
{\bf 32.4} % Context
{\bf 34.8} % Object
{\underline{62.1}} % VOC
\\ %ADE 

\bottomrule
\end{tabular}   

\vspace{-5pt}

\caption{\textbf{Open-vocabulary semantic segmentation quantitative comparison} using the mIoU metric. We separate the datasets used for evaluation into two columns: 
those without a `background' prompt and those with (noted `W/ bkg prompt'), as discussed in \cref{sec:setup}. 
We report all methods without post-processing. We note with $^*$ methods for which we computed scores; we obtained MaskCLIP$^{*}$ scores with OpenCLIP~\cite{openclip} and mark with $\dag$ the use of MaskCLIP refinement. 
The first and second best methods are respectively \textbf{bold} and \underline{underlined}.
We specify if a method assumes prior access to names of concepts (`Concept spec.') and what additional data is used at training (`Extra data').
`IN' stands for ImageNet \cite{imagenet} and `SD' for Stable Diffusion \cite{rombach2022high}.
We refer to \cref{sec:setup} for more details
on baselines.
}
\label{table:main_nobg}
\label{table:main_bg}
\vspace{-10pt}
\end{table*}

\subsection{Open vocabulary semantic segmentation}
\label{sec:zero-shot}
We discuss in this section state-of-the-art results on the task of open-vocabulary semantic segmentation. 

\parag{Evaluation with no `background' class.} 
We first compare in \cref{table:main_nobg} (`No background prompt' column) the results on datasets which aim at the segmentation of most of the pixels in an image and do not consider a `background' class. We observe that our method \ours achieves the best results on four datasets yielding  +$2.2$, +$5.0$, +$6.7$ and +$5.1$ mIoU over the second best performing method. Interestingly, we outperform methods which build expensive prototypes per visual concept on fine-grained datasets, showing the benefit of our lightweight and generalizable method.
The only drop (-$0.8$ mIoU) is seen on VOC20 with respect to OVDiff; we believe it is due to the benefit of generating per-concept negative prototypes which likely benefits this object-centric dataset. 
An adaptive granularity of feature correlation could help mitigate this drop, which we leave for future work.

\begin{figure*}[ht!]
\centering
\renewcommand{\arraystretch}{0}
\setlength\tabcolsep{0pt} 

\resizebox{\textwidth}{!}{
\begin{tabular}{
>{\centering\arraybackslash}m{1em}
>{\centering\arraybackslash}m{0.13\linewidth}
>{\centering\arraybackslash}m{0.13\linewidth}
>{\centering\arraybackslash}m{0.13\linewidth}
>{\centering\arraybackslash}m{0.13\linewidth}
>{\centering\arraybackslash}m{0.13\linewidth}
>{\centering\arraybackslash}m{0.13\linewidth}
>{\centering\arraybackslash}m{6em}
}
& RGB & GT & {TCL~\cite{cha2022tcl}} &  {CLIP-DIY~\cite{wysoczanska2023clipdiy}} & \small{MaskCLIP} & \small{\ours} & \small{prompts} \\
\raisebox{-.1\normalbaselineskip}[0pt][0pt]{\rotatebox[origin=c]{90}{\small{VOC}}}
& \includegraphics[width=\linewidth, height=1.5cm]{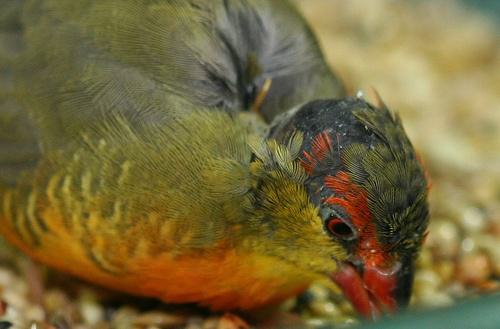} & \includegraphics[width=\linewidth, height=1.5cm]{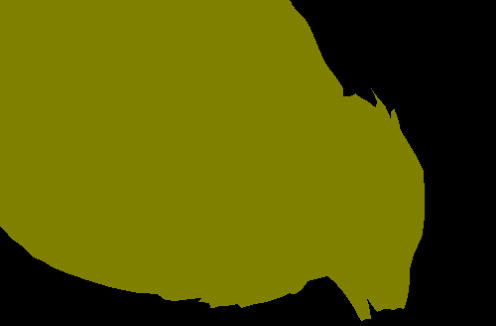} & \includegraphics[width=\linewidth, height=1.5cm]{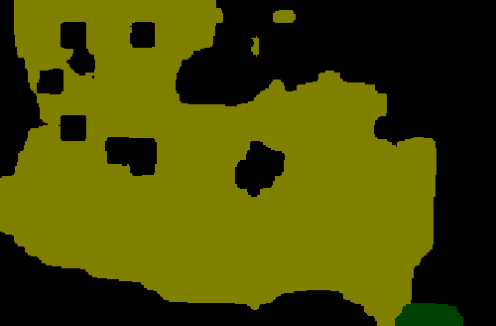} & \includegraphics[width=\linewidth, height=1.5cm]{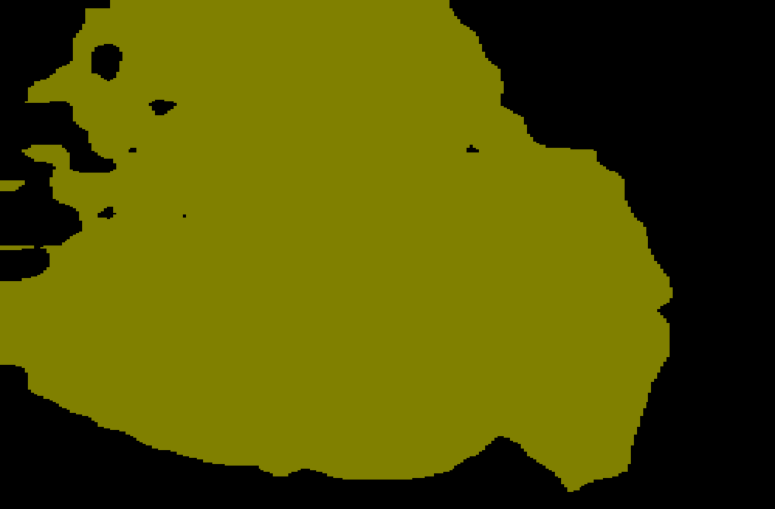} & \includegraphics[width=\linewidth, height=1.5cm]{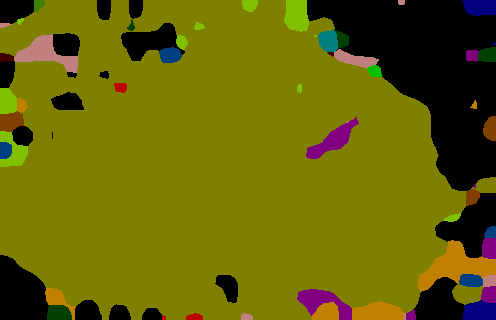} & \includegraphics[width=\linewidth, height=1.5cm]{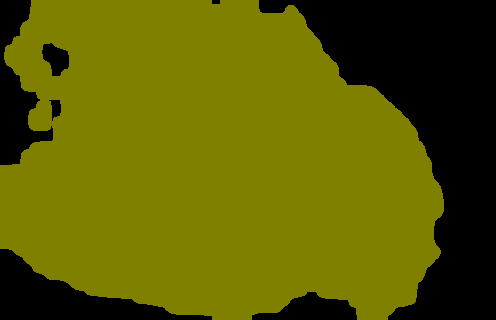} & {\cellcolor[gray]{.9} 
 \makecell{\promptstyle{plant_pascal}{potted plant} \\ \promptstyle{bird_pascal}{bird}}} \\

% \rotatebox[origin=c]{90}{\small Context59} 
\raisebox{-.2\normalbaselineskip}[0pt][0pt]{\rotatebox[origin=c]{90}{\small{Context59}}} & 
\includegraphics[width=\linewidth, height=1.5cm]{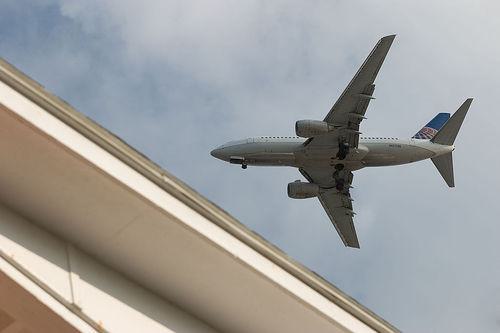} & 
\includegraphics[width=\linewidth, height=1.5cm]{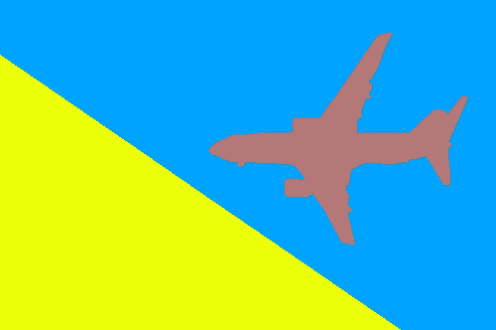} & 
\includegraphics[width=\linewidth, height=1.5cm]{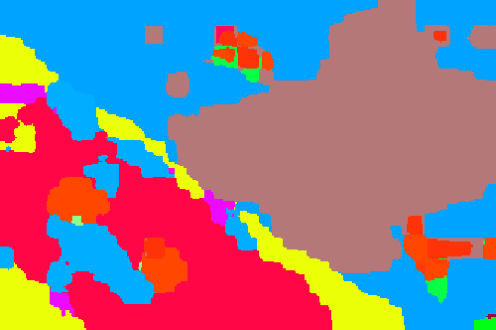} & 
\includegraphics[width=\linewidth, height=1.5cm]{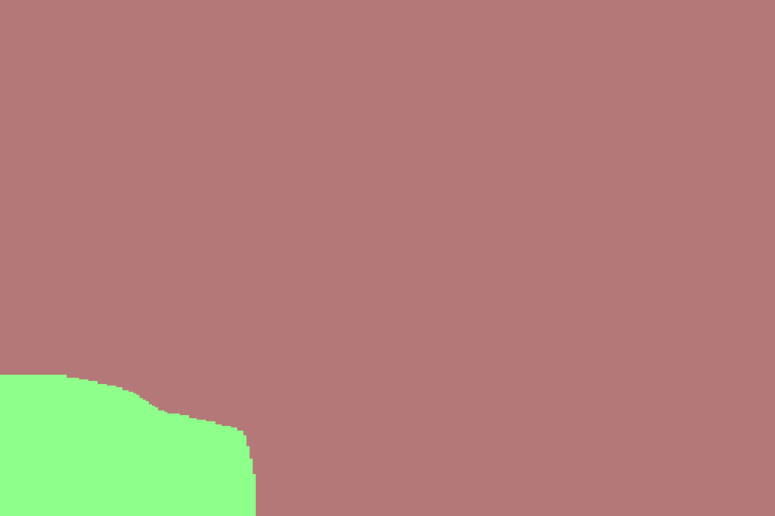} &
\includegraphics[width=\linewidth, height=1.5cm]{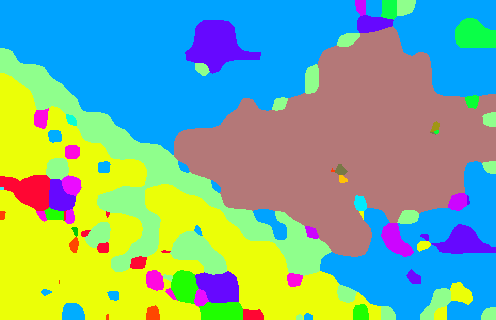} & 
\includegraphics[width=\linewidth, height=1.5cm]{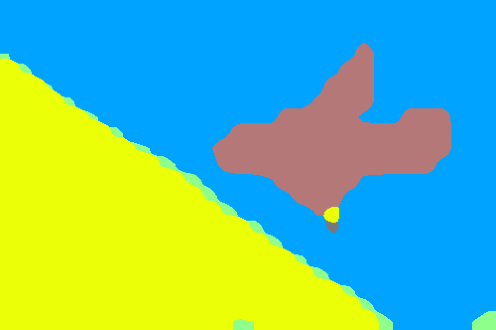} & 
{\cellcolor[gray]{.9} 
 \makecell{\promptstyle{aeroplane_context}{aeroplane} \\  \promptstyle{sky_context}{sky} \promptstyle{building_context}{building} \\ \promptstyle{door_context}{door} }} \\
 
% \rotatebox[origin=c]{90}{\small COCO}  
\raisebox{-.2\normalbaselineskip}[0pt][0pt]{\rotatebox[origin=c]{90}{\small{COCO}}} & 
\includegraphics[width=\linewidth, height=1.5cm]{figures/images/000000273642.jpg} &  
\includegraphics[width=\linewidth, height=1.5cm]{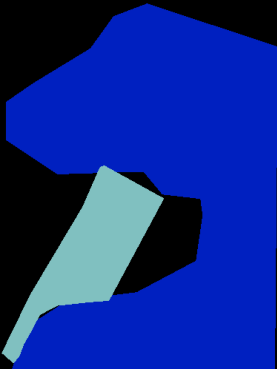} & 
\includegraphics[width=\linewidth, height=1.5cm]{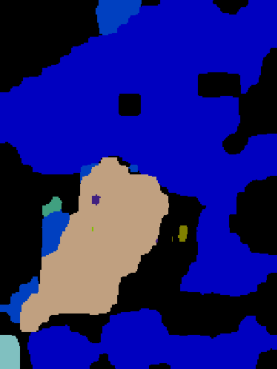} & 
\includegraphics[width=\linewidth, height=1.5cm]{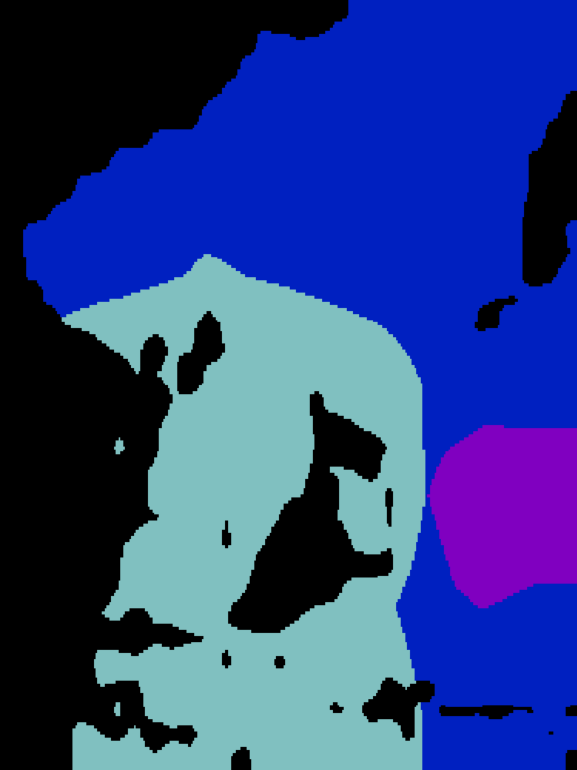} & 
\includegraphics[width=\linewidth, height=1.5cm]{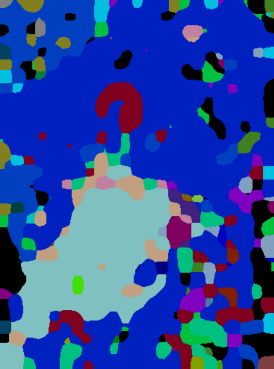} & 
\includegraphics[width=\linewidth, height=1.5cm]{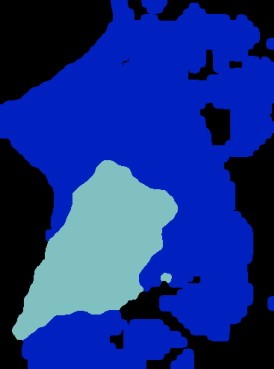} & 
{\cellcolor[gray]{.9}\makecell
{\promptstyle{dog_coco}{dog} \promptstyle{remote}{remote}  \\ \promptstyle{cell phone_coco}{cell phone}}} \\

% \rotatebox[origin=c]{90}{\small Cityscapes}  
\raisebox{-.2\normalbaselineskip}[0pt][0pt]{\rotatebox[origin=c]{90}{\small{Cityscapes}}} & 
\includegraphics[width=\linewidth, height=1.5cm]{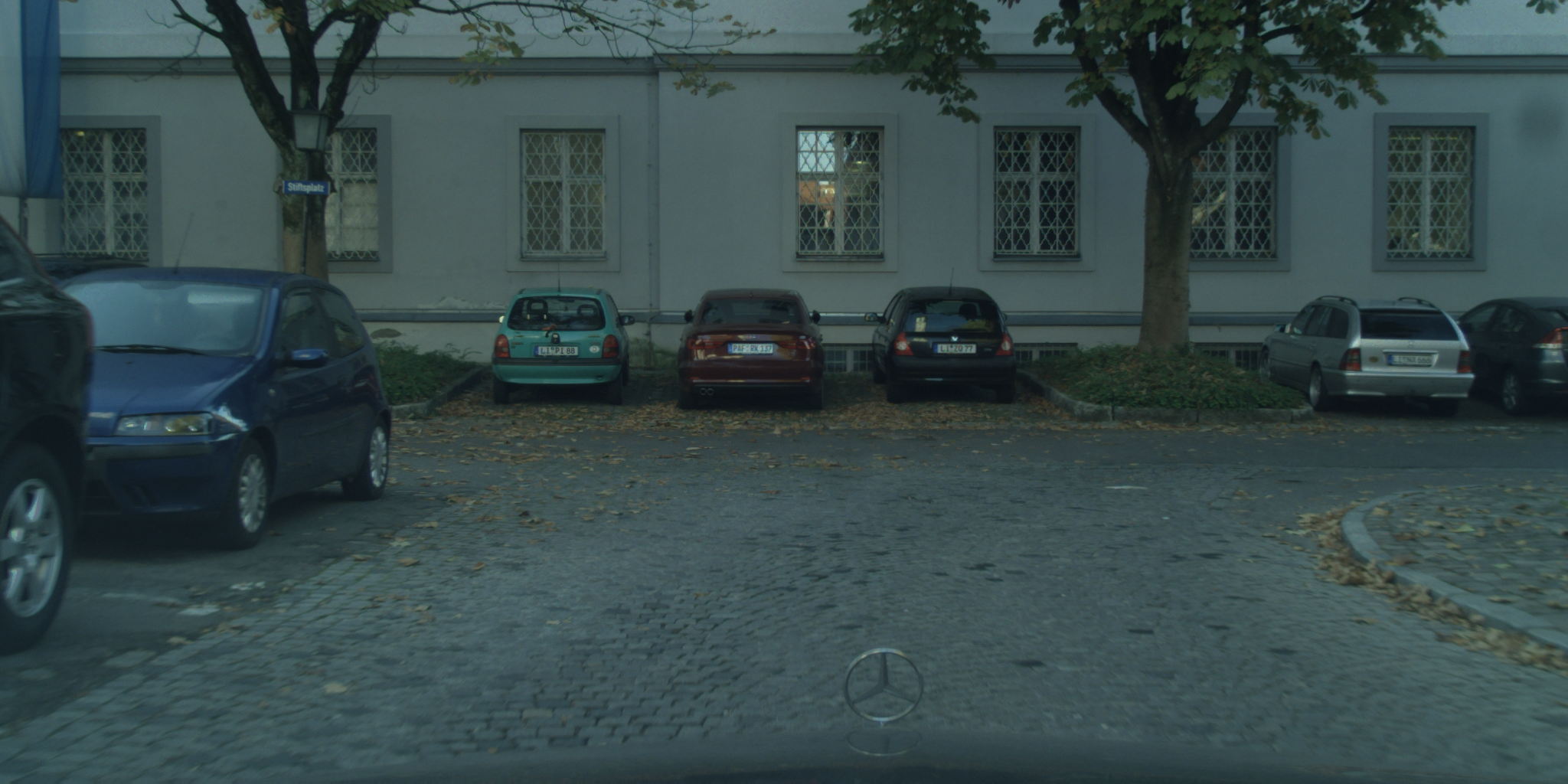} & 
\includegraphics[width=\linewidth, height=1.5cm]{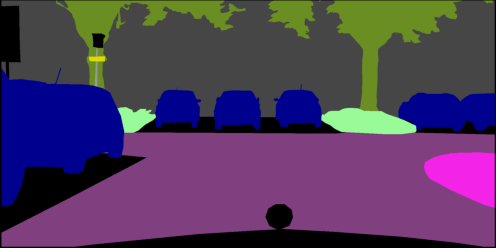} &
\includegraphics[width=\linewidth, height=1.5cm]{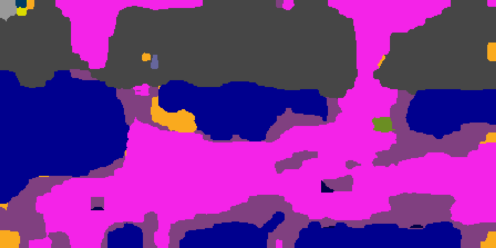}  & 
\includegraphics[width=\linewidth, height=1.5cm]{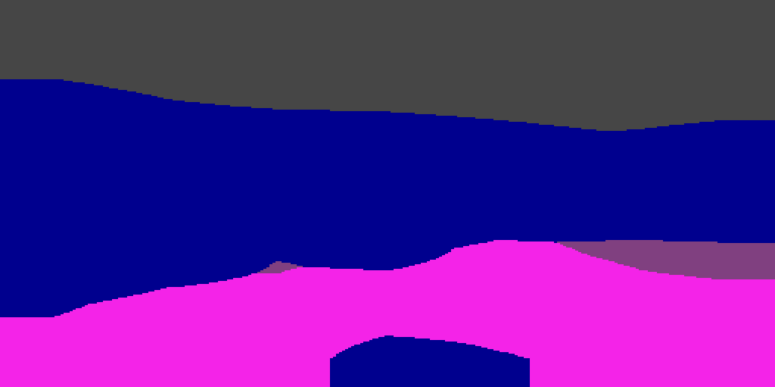}  & 
\includegraphics[width=\linewidth, height=1.5cm]{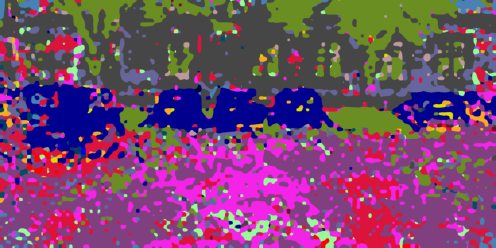} & 
\includegraphics[width=\linewidth, height=1.5cm]{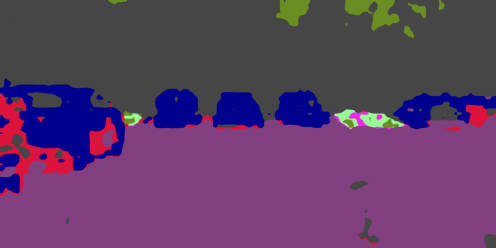}  & 
{\cellcolor[gray]{.9}\makecell 
{\promptstyle{road_city}{road} \promptstyle{car_city}{car} \\ \promptstyle{sidewalk_city}{sidewalk} \\ \promptstyle{vegetation_city}{vegetation}}}
\\

\raisebox{-.2\normalbaselineskip}[1pt][1pt]{
\rotatebox[origin=c]{90}{ADE}}& 
\includegraphics[width=\linewidth, height=1.5cm]{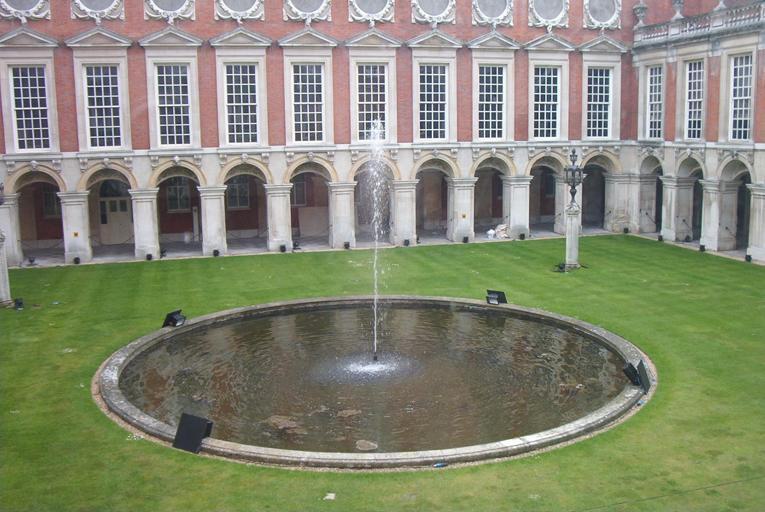} & 
\includegraphics[width=\linewidth, height=1.5cm]{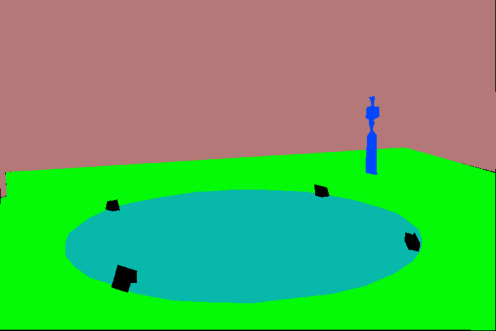} & 
\includegraphics[width=\linewidth, height=1.5cm]{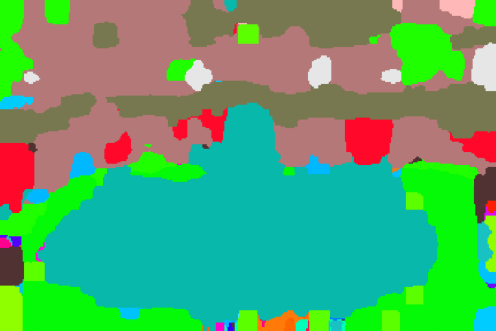} & 
\includegraphics[width=\linewidth, height=1.5cm]{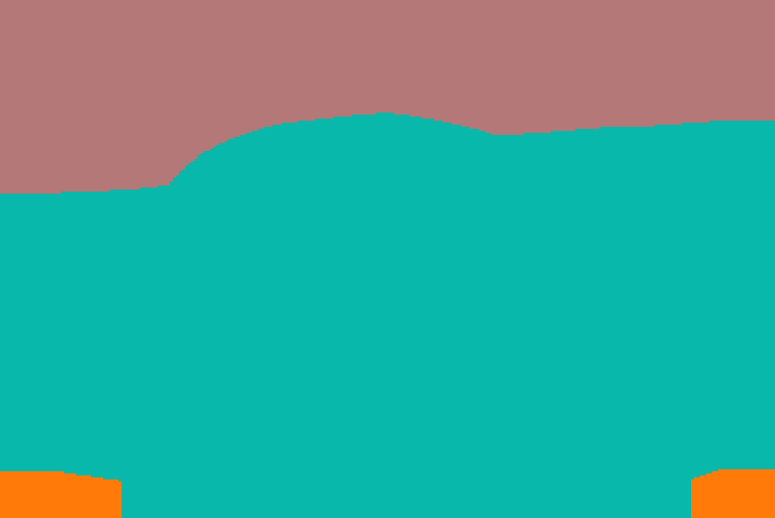} & 
\includegraphics[width=\linewidth, height=1.5cm]{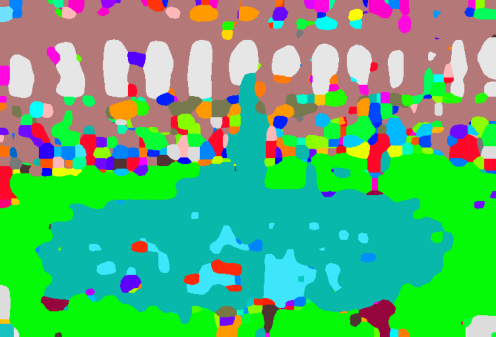} & 
\includegraphics[width=\linewidth, height=1.5cm]{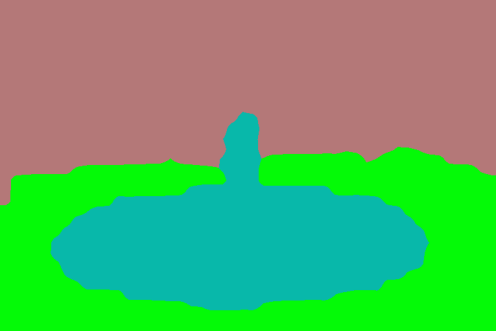} & 
{\cellcolor[gray]{.9}\makecell{
\promptstyle{fountain_ade}{fountain}  \\ \promptstyle{building_ade}{building} \\ \promptstyle{grass_ade}{grass}} }\\

\end{tabular}
}

\vspace{-6pt}
\caption{
\textbf{Qualitative open-vocabulary segmentation results}. We compare ours against CLIP-DIY~\cite{wysoczanska2023clipdiy},  TCL~\cite{cha2022tcl} and MaskCLIP~\cite{zhou2022maskclip}. For a fair comparison, we do not apply post-processing. 
All pixels annotated in black are from the background class. 
}
\vspace{-10pt}
\label{fig:results_main}
\end{figure*}

\parag{Evaluation with `background' class.} We now compare our method on datasets which include a `background' query in~\cref{table:main_bg} (`W/ bkg prompt' column).  
In this setup, we also apply our background detection mechanism (detailed in \cref{sec:background}) on VOC and Object in order to improve the stuff-like background detection. 
We observe that \ours significantly outperforms all methods which do not construct prototypes. Moreover, we surpass OVDiff (which uses an ensemble of three models) on Context dataset by +$2.3$ mIoU and are on par on Object. 
It is to be noted that with a single feature extractor, the performance of OVDiff drops by 
-$10$ mIoU and the method requires the construction of a `background' prototype \emph{per concept}, otherwise losing another -$10$ mIoU on VOC. On the other hand, \ours produces segmentation masks in a \emph{single} pass of CLIP with the light addition of two convolutional layers while remaining fully open-vocabulary as it does not require \emph{any} concept-specific constructs.

\parag{Qualitative results.} We qualitatively compare in \cref{fig:results_main} \ours with high-performing TCL~\cite{cha2022tcl}, CLIP-DIY~\cite{wysoczanska2023clipdiy} (two recent methods which provide code) and our baseline method MaskCLIP~\cite{zhou2022maskclip} on images taken from the datasets considered in the evaluation. We observe that our method generates predictions accurate both in terms of localization and assignment. Indeed we obtain fined-grained results on the challenging datasets, e.g. in the Cityscapes example the text query `car' and in the ADE20k example `fountain' are accurately located when CLIP-DIY and TCL produce coarser results. Versus MaskCLIP, we can see the denoising capabilities of \ours as MaskCLIP hallucinations grow with the number of text queries prompted at evaluation.
Finally, in Fig.~\ref{fig:first} we present 'in the wild' examples, beyond the evaluation benchmarks, and show that \ours produces accurate segmentation masks for arbitrary and very specific prompts, such as `wooden table' or `leather bag'.

\subsection{Ablation study}
\label{sec:ablation}

We now conduct an ablation study of 
the different 
components of 
\ours and investigate the impact of both our feature pooling strategy and background detection. 

\parag{The impact of the pooling mechanism.} We propose with \ours to combine MaskCLIP \emph{features} with a well-defined linear combination and compare different solutions in \cref{tab:abl-pooling}. 
In~\cite{zhou2022maskclip}, the authors proposed to refine the \emph{predictions} with a combination weighted by CLIP \emph{key} embeddings (noted `CLIP keys (preds.)' in the table) and boost MaskCLIP results by more than +$8$ mIoU on VOC and VOC20, +$1.8$ and +$1.0$ and +$0.6$ mIoU on the other datasets. However, we show that working directly at the feature level allows us to achieve better results; we obtain consistent improvements ranging from +$6$ to +$19$ mIoU on all datasets when using DINO-based weight $\dinocorr$ and further improve when using trained CLIP-based weights $\clipcorr$. 

\begin{table}[th!]
\centering
\setlength{\tabcolsep}{1.8pt}
\begin{subtable}[t]{\linewidth}
    \centering
    \resizebox{0.95\linewidth}{!}{
    \begin{tabular}{l|ccccc}
        \toprule
         Pooling strategy & VOC & VOC20 & C59 & Stuff & ADE \\ 
        \midrule
        \rowcolor{white!90!black} MaskCLIP 
        \cite{zhou2022maskclip} \textit{- baseline} & 32.9 & 61.8 & 25.6& 17.6 & 14.3\\

        CLIP keys (preds.) \cite{zhou2022maskclip} & 41.3 &	71.9 & 27.4	& 18.6	& 14.9 \\
        ours w. CLIP keys & 39.2	& 73.2 & 23.0 & 12.6	& 7.7\\
        ours w. DINO $\dinocorr$ & 53.7 & 79.1 & 35.5 &	24.7 & 20.4 \\
        \rowcolor{ForestGreen!15!white} ours w. trained $\clipcorr$ & 54.0 & 80.9 & 35.9 &  24.6 &  20.0 \\
        \bottomrule
    \end{tabular}
    }
    \caption{\textbf{Pooling strategy}}
    \label{tab:abl-pooling}
\end{subtable}

\begin{subtable}[t]{\linewidth} 
    \centering
    \resizebox{0.75\linewidth}{!}{
    \begin{tabular}{l|l|cc}
        \toprule
         Pooling  & Bkg det. & Object & VOC \\ 
        \midrule
        \rowcolor{white!90!black} \multicolumn{2}{l}{MaskCLIP~\cite{zhou2022maskclip} \textit{- baseline}} & 16.4 & 32.9 \\
        ours w. DINO $\dinocorr$ &  & 29.9 & 53.7\\
        ours w. DINO $\dinocorr$ &  FOUND & 32.1 & 60.1  \\
        ours w. DINO $\dinocorr$ & ours w. $M$ & 34.1 & 62.1\\
        ours w. DINO $\dinocorr$ & ours w. $M^\phi$ & 34.2 & 61.9 \\
        \rowcolor{ForestGreen!15!white} ours w. trained $\clipcorr$ & ours w. $M^\phi$ 	& 34.8 & 62.1\\
        \bottomrule
    \end{tabular}
    }
    \caption{\textbf{Background detection}}
    \label{tab:abl-bkg}
\end{subtable}
\caption{\textbf{Impact of the pooling strategy} (a) and background detection (b) on diverse datasets reported with the mIoU metric.}
\end{table} 

\parag{The impact of the background detection.} We now discuss the improvement provided by our background refinement strategy, which is applied when \emph{stuff}-like background patches need to be detected.
We report such results in \cref{tab:abl-bkg} when employing our pooling strategy (either using DINO features, noted `w. DINO $\dinocorr$' or those extracted from CLIP, noted `w. trained $\clipcorr$'). When using solely `FOUND' for background detection, as in~\cite{wysoczanska2023clipdiy}, we improve by +$6.4$ mIoU on VOC (achieving $60.1$ mIoU), but when relaxing FOUND (see \cref{sec:background}) with an uncertainty condition, we boost scores up to 62.1 on VOC, showing the limitation of using FOUND alone. We also achieve similar results when using CLIP-based predictions $M^\phi$ both with DINO-based $\dinocorr$ and trained CLIP-based $\clipcorr$ correlations, although we observe that best results are overall obtained with trained $\clipcorr$.
We visualize CLIP-based mask $M^\phi$ in \cref{fig:found-pred} 
and see high similarity to DINO-based predictions, therefore showing the localization quality of CLIP.

\vspace{-6pt}
\section{Conclusions}
\vspace{-4pt}
In this work, we propose to make the most out of CLIP features and show that the features already contain useful \emph{localization information}. Indeed with light convolutional layers, we are able to learn both good patch-correlation and objectness information by using DINO self-supervised model as a guide. With such information, our method \ours 
performs zero-shot open-vocabulary semantic segmentation in a single pass of CLIP model and with two light extra convolutional layers. \ours reaches state-of-the-art results on complex semantic segmentation datasets. 
 
\parag{Limitations.} Despite yielding strong results on 
open-vocabulary semantic segmentation, \ours is still bounded by the capability of the CLIP model to separate classes, as it inherits its granularity. We believe that better prompt engineering paired with better image-text models could further boost the performance of \ours.

\section*{Acknowledgments}
This work was supported by the National Centre of Science (Poland) Grant No. 2022/45/B/ST6/02817 and by the grant from NVIDIA providing one RTX A5000 24GB used for this project.
% \newpage
{
    \small
    \bibliographystyle{ieeenat_fullname}
    \bibliography{main}
}
\clearpage
% \maketitlesupplementary
\setcounter{section}{0}
\setcounter{table}{2}
\setcounter{figure}{7}
\renewcommand{\thesection}{\Alph{section}}
\def\graystrength{0.75}

% \columnbreak
\vfill\null

\newcommand{\std}[1] {\scriptsize{\textcolor{ForestGreen}{\textbf{$\pm$#1}}}}

\section{More experimental results}
\subsection{The impact of the training dataset}
\label{sec:sup_exp}
\parag{Training stability.} 
We report in the main paper the final results averaged over three different randomly sampled subsets of ImageNet used for the training. In the first row of~\cref{tab:datasets} we report the corresponding standard deviation. We observe that in all cases the standard deviation equals $0.1$ mIoU or lower, therefore showing the stability of our training.

\parag{Training with different datasets.} Our method \ours does not require any labels to be trained. We investigate here the impact of training on the datasets used to train self-supervised DINO \cite{caron2021dino} and FOUND \cite{simeoni2023found}, namely Imagenet and DUTS-TR~\cite{wang2017duts}. We report scores in \cref{tab:datasets}. We also provide results when increasing the dataset size to 10k on ImageNet. 
In all cases, we observe no significant difference when using one dataset or another, and the size of the dataset does not seem to impact results positively. 

\begin{table}[h!]
    \centering
    \newcommand{\resultsrowbg}[4]{#1 & #2 & #3 & #4}
\newcommand{\resultsrowbgtwo}[6]{#1 & #2 & #3 & #4 & #5 & #6}

\begin{subtable}[t]{\linewidth}
    \centering
    \setlength{\tabcolsep}{1.5pt}
    \begin{tabular}{l|lllll}
    \toprule
    \resultsrowbgtwo{{Train. dataset}}
    {C59}
    {V20}
    {Stuff}
    {City}
    {ADE}
    \\
    \midrule
    
    \resultsrowbgtwo{{IN-1k}}
    {35.9\std{0.1}} % VOC20
    {80.9\std{0.0}} % Context59
    {24.6\std{0.1}} % Stuff
    {31.7\std{0.1}} % City
    {20.0\std{0.0}} %ADE 
    \\
    
    \resultsrowbgtwo{{IN-10k}}
    {35.9\std{0.0}} % VOC20
    {80.3\std{0.1}} % Context59
    {24.7\std{0.0}} % Stuff
    {31.9\std{0.1}} % City
    {20.1\std{0.0}} %ADE 
    \\
    
    % \resultsrowbg{DUT}
    \resultsrowbgtwo{DUTS-TR~\cite{wang2017duts}}
    {35.9} % VOC20
    {80.5} % Context59
    {24.6} % Stuff
    {31.3} % City
    {19.9} %ADE 
    \\
    \bottomrule
    \end{tabular}
    % \caption{\textbf{Without bkg prompt}}
    \label{tab:sup_bkg}
    \caption{Benchmark \textbf{without} `background' prompt}
\end{subtable}

\vspace{10pt}

\begin{subtable}[t]{\linewidth}
\centering
    \setlength{\tabcolsep}{1.5pt}
    \label{tab:sup_nobkg}
    % \resizebox{\linewidth}{!}{
    \begin{tabular}{l|lll}
    \toprule
    \resultsrowbg{{Train. dataset}}
    {{VOC}}  %{VOC}}
    {Con.}
    {Obj}
    \\
    \midrule
    
    \resultsrowbg{{IN-1k}}
    {62.1\std{0.0}} % VOC
    {32.4\std{0.1}} % Context
    {34.8\std{0.1}} % Object
    \\
    
    \resultsrowbg{{IN-10k}}
    {61.9\std{0.0}} % VOC
    {32.4\std{0.0}} % Context
    {34.6\std{0.1}} % Object
    \\
    
    \resultsrowbg{DUTS-TR~\cite{wang2017duts}}
    % \resultsrowbg{DUTS-TR~\cite{wang2017duts}}
    {62.0} % VOC
    {32.4} % Context
    {34.8} % Object
    \\
    \bottomrule
    \end{tabular}
    % }
    \caption{Benchmark \textbf{with} `background' prompt}
\end{subtable}% <---- don't forget this %

    \caption{\textbf{Performance with different training datasets.} When using random splits extracted from ImageNet (noted `IN'), we report the average score and \textcolor{ForestGreen}{standard deviation} computed over training with three random splits (of 1k or 10k) extracted in ImageNet. In (a) we report the scores on the datasets without 'background' class and in (b) with.} 
    \label{tab:datasets}
\end{table}

\begin{figure}
    \renewcommand{\arraystretch}{0.2}
    \setlength\tabcolsep{0.5pt} 
    \centering
    \subfloat[][\textbf{Input} image]{\includegraphics[width=0.45\linewidth]{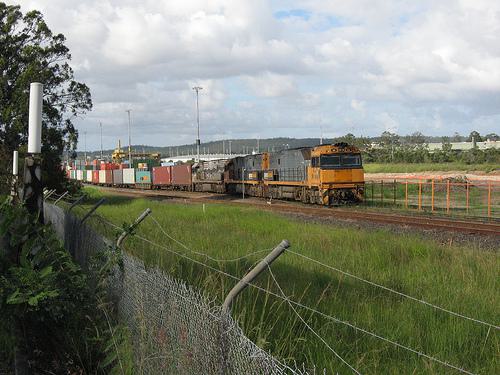}\label{fig:sup-correlation-image}} \\ 
    \subfloat[][\textbf{Correlation maps} for different seeds (in \textbf{\color{red}red})]{
        \begin{tabular}{ccc}
           \emph{query} & \emph{key} & \emph{value} \\
        \includegraphics[width=0.25\linewidth]{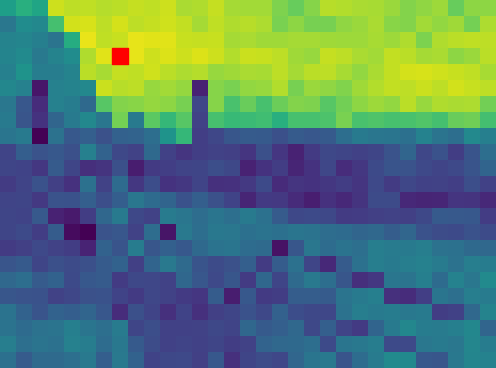} &  \includegraphics[width=0.25\linewidth]{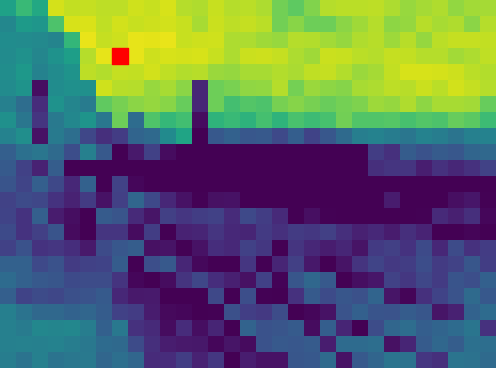} &  \includegraphics[width=0.25\linewidth]{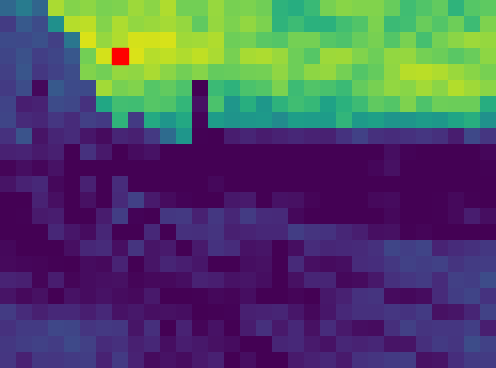} \\
          \includegraphics[width=0.25\linewidth]{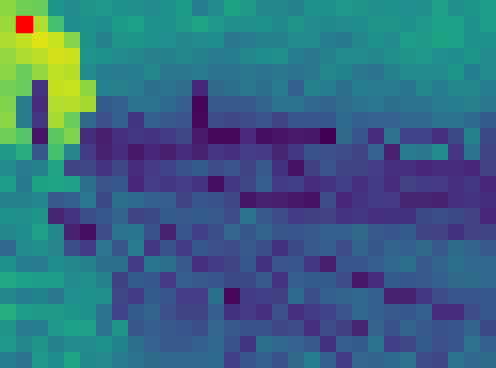}   &  \includegraphics[width=0.25\linewidth]{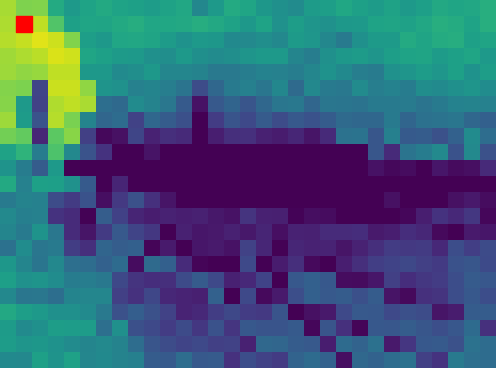} &  \includegraphics[width=0.25\linewidth]{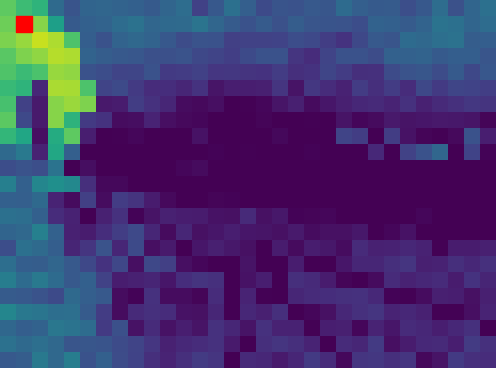}
           \\
          \includegraphics[width=0.25\linewidth]{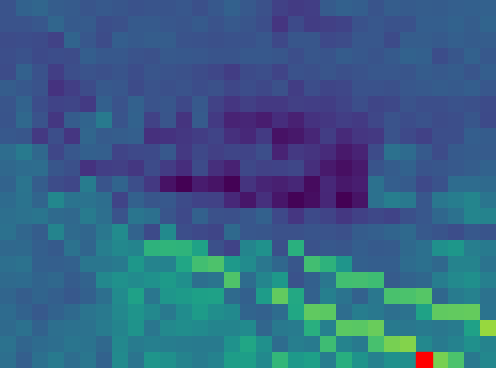}   &  \includegraphics[width=0.25\linewidth]{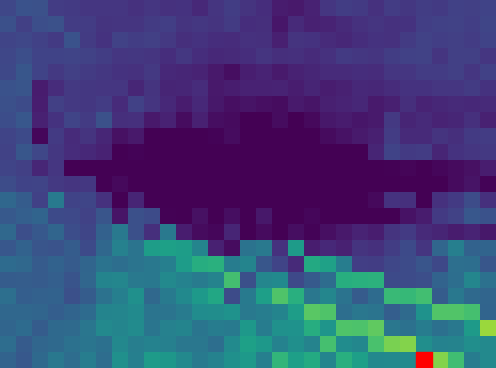} &  \includegraphics[width=0.25\linewidth]{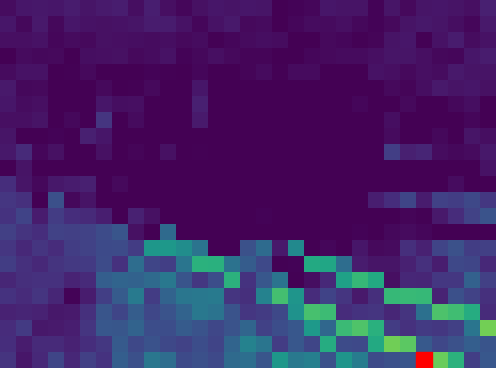}
        \end{tabular}
        \label{fig:sup-correlation-kqv}
    }
    \\
    \subfloat[][Resulting \textbf{segmentation} maps]{
        \centering
        \begin{tabular}{
        >{\centering\arraybackslash}m{0.27\linewidth}
        >{\centering\arraybackslash}m{0.27\linewidth}
        >{\centering\arraybackslash}m{0.27\linewidth}
        }
           \emph{query} & \emph{key} & \emph{value} \\
        \includegraphics[width=\linewidth]{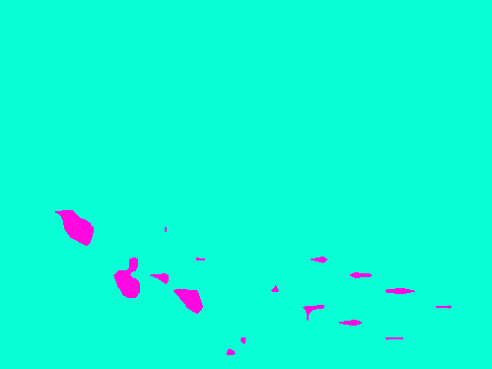} &  \includegraphics[width=\linewidth]{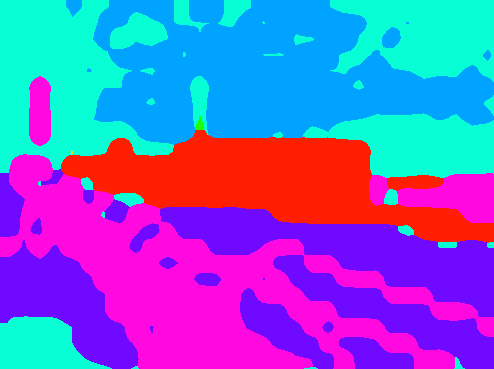} &  \includegraphics[width=\linewidth]{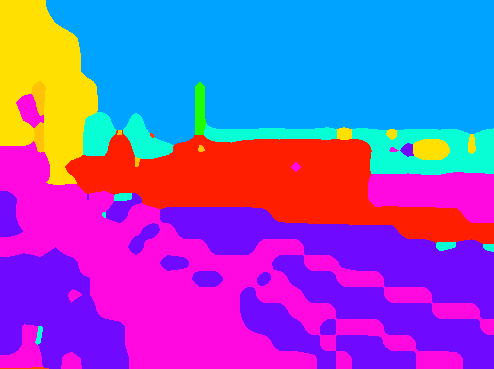} \\
        \multicolumn{3}{>{\centering\arraybackslash}m{0.82\linewidth}}{{\cellcolor[gray]{\graystrength} 
             \makecell{\promptstyle{sky_context}{sky} \promptstyle{tree_context}{tree} \promptstyle{train_context}{train} \promptstyle{ground_context}{ground} \promptstyle{fence_context}{fence} \promptstyle{grass_context}{grass}}}} \\
        \end{tabular}
        \label{fig:sup-correlation-seg}
    }
    \\
    \caption{\textbf{Visualization of correlation and segmentation} obtained with different embeddings of DINO: \emph{query}, \emph{key} and \emph{value}.
    }
    \label{fig:sup-correlation}
\end{figure}

\subsection{Self-supervised features discussion}
\label{sec:self-abl}
We present in \cref{fig:sup-correlation} visualizations of correlation obtained using different DINO embeddings extracted from DINO's last attention layer, namely `query', `key' and `value'. Most unsupervised localization methods \cite{simeoni2021lost, wang2022tokencut, simeoni2023found, wang2023cut} use the `key' embeddings which allow the easy separation of \emph{foreground} from \emph{background}. However, we observed in this work that using instead the \emph{value} features allows us to separate better elements in the background, as visible in the figure. Patches in the background correlate to fewer background patches and regions are therefore better separated.

We also depict the final segmentation when using each type of feature, and observe the best result with `value'. We observe that more objects in the background are well-segmented and labeled, e.g., `tree' and `sky'.

\subsection{Background evaluation with FOUND}
\begin{table}
    \setlength{\tabcolsep}{2pt}
     \resizebox{\columnwidth}{!}{
    \centering
    \begin{tabular}{l|ccc|ccc}
    \toprule
     & \multicolumn{3}{c}{\textbf{Single obj. discovery}} &  \multicolumn{3}{c}{\textbf{Unsup. saliency detection}}  \\ \\
     % \midrule
         Method & VOC7 & VOC12 & C20k & DUT-O. & DUTS-T. & ECSSD \\
         \midrule
         FOUND \cite{simeoni2023found} & 72.5 & \bf 76.1 & 62.9 & \bf60.8 &  65.4 & 80.5 \\
         ours & \bf 73.1 &  75.9 & \bf 64.4 & 60.6 & \bf 66.6 & \bf 81.3 \\ \bottomrule
    \end{tabular}
    }
    \caption{Results of \textbf{single object discovery and unsupervised saliency detection} obtained when following FOUND \cite{simeoni2023found} protocol. We compute the single object discovery scores on classic VOC benchmarks \cite{pascal-voc-2012} and 20k images of COCO (noted `C20k') following \cite{simeoni2023found} and use the CorLoc metric. We report the mIoU metric for unsupervised saliency detection and provide all results with the post-processing bilateral solver. We note `DUT-O.' DUT-OMRON \cite{yang2013saliency} and `DUTS-T.' stands for DUTS-TEST \cite{wang2017duts}.}
    \label{fig:sup:objdiscovery}
\end{table}

\newlength{\ablcocowidth}
\setlength{\ablcocowidth}{0.29\linewidth}

\newlength{\ablcocoheight}
\setlength{\ablcocoheight}{2cm}

\begin{figure}[h]
\centering
\renewcommand{\arraystretch}{0}
\setlength\tabcolsep{0pt} 

\begin{center}
\begin{tabular}{
>{\centering\arraybackslash}m{2.8em} 
>{\centering\arraybackslash}m{\ablcocowidth}
>{\centering\arraybackslash}m{\ablcocowidth}
>{\centering\arraybackslash}m{\ablcocowidth}
}

RGB & \addcocoablresultjpeg{\ablcocowidth}{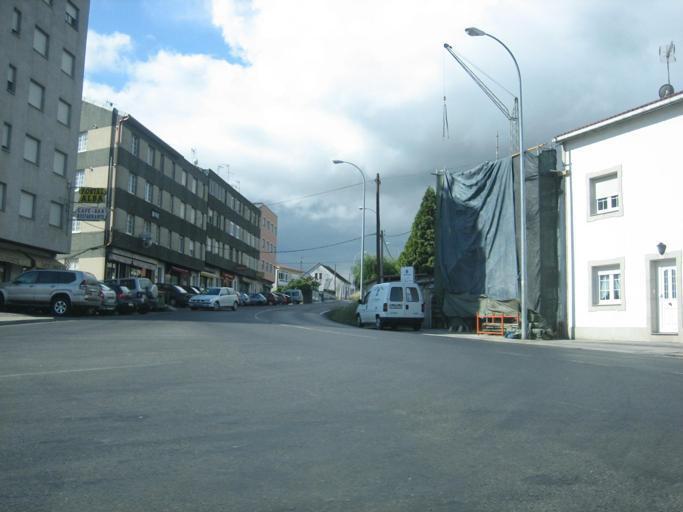} &
\addcocoablresultjpeg{\ablcocowidth}{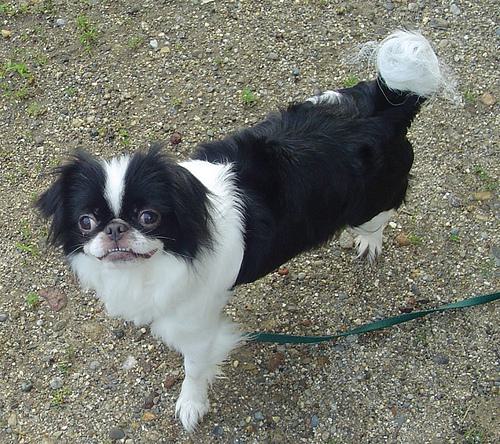} &
\addcocoablresult{\ablcocowidth}{frankfurt_000000_003025}{} \\

GT & \addcocoablresult{\ablcocowidth}{ADE_val_00000813}{gt} &
\addcocoablresult{\ablcocowidth}{2009_000181}{gt} &
\addcocoablresult{\ablcocowidth}{frankfurt_000000_003025}{gt} \\
 % {ADE} &

Mask CLIP & \addcocoablresult{\ablcocowidth}{ADE_val_00000813}{maskclip} &
\addcocoablresult{\ablcocowidth}{2009_000181}{maskclip} &
\addcocoablresult{\ablcocowidth}{frankfurt_000000_003025}{maskclip} \\

Ours & \addcocoablresult{\ablcocowidth}{ADE_val_00000813}{ours} &
\addcocoablresult{\ablcocowidth}{2009_000181}{ours} &
\addcocoablresult{\ablcocowidth}{frankfurt_000000_003025}{ours} \\

Class. & {\cellcolor[gray]{\graystrength}\makecell
{\promptstyle{building_ade}{building} \\ \promptstyle{car_ade}{car} \promptstyle{sky_ade}{sky} \promptstyle{road_ade}{road}\\
 \promptstyle{house_ade}{house}  \promptstyle{washer_ade}{washer} \\\promptstyle{sidewalk_ade}{sidewalk} }} 
& {\cellcolor[gray]{\graystrength}\makecell
{\promptstyle{dog_context}{dog}  \promptstyle{road_context}{road} \\ \promptstyle{ground_context}{ground} \\
 \promptstyle{door_context}{door} \promptstyle{grass_context}{grass} \\
 \promptstyle{floor_context}{floor}
 }}
& 
{\cellcolor[gray]{\graystrength}\makecell
{\promptstyle{road_city}{road} \promptstyle{person_city}{person} \\  \promptstyle{car_city}{car} \promptstyle{sidewalk_city}{sidewalk} \\ 
\promptstyle{bicycle_city}{bicycle} \\
\promptstyle{vegetation_city}{vegetation} 
}} \\
\end{tabular}
\end{center}

    \caption{
    \textbf{Visual ablations of the impact of our pooling method}. Examples from ADE20K (left), PASCAL Context (middle), and Cityscapes (right) datasets. 
    }
    \label{fig:sup_maskclip_abl}
\end{figure}
\newlength{\resultsheight}

\newcommand{\addresulttwo}[3]{%
\begin{minipage}[b]{#1}
     \centering
     \includegraphics[width=\textwidth, height=\resultsheight]{figures/images/#2_#3.png}
\end{minipage}}
\newcommand{\addresulttwojpeg}[2]{%
\begin{minipage}[b]{#1}
     \centering
     \includegraphics[width=\textwidth, height=\resultsheight]{figures/images/#2.jpg}
\end{minipage}}

\setlength{\ablcocowidth}{0.135\textwidth}

\setlength{\resultsheight}{1.8cm}

\begin{figure}[h!]
\centering
\renewcommand{\arraystretch}{0}
\setlength\tabcolsep{0pt} 

\begin{tabular}{
>{\centering\arraybackslash}m{3.em} 
>{\centering\arraybackslash}m{\ablcocowidth}
>{\centering\arraybackslash}m{\ablcocowidth}
>{\centering\arraybackslash}m{\ablcocowidth}
}
RGB & \addresulttwojpeg{\ablcocowidth}{000000207728}  & 
\addresulttwojpeg{\ablcocowidth}{000000148957} 
&
\addresulttwojpeg{\ablcocowidth}{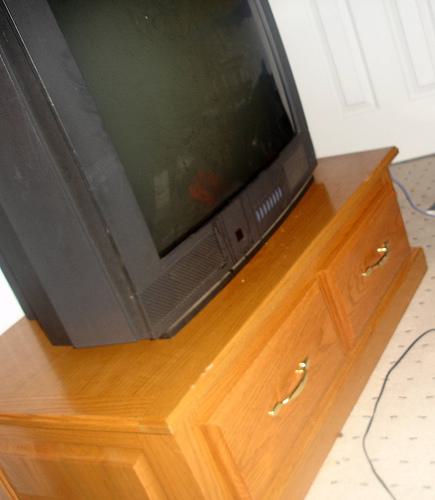}\\

GT & \addresulttwo{\ablcocowidth}{000000207728}{gt}  &
\addresulttwo{\ablcocowidth}{000000148957}{gt} 
& \addresulttwo{\ablcocowidth}{2008_002650}{gt}\\

Ours with bkg & \addresulttwo{\ablcocowidth}{000000207728}{oursnofound} &
\addresulttwo{\ablcocowidth}{000000148957}{oursnofound}  &
\addresulttwo{\ablcocowidth}{2008_002650}{oursnofound}\\

Ours w/o bkg & 
\addresulttwo{\ablcocowidth}{000000207728}{ours} &
\addresulttwo{\ablcocowidth}{000000148957}{ours}  &
\addresulttwo{\ablcocowidth}{2008_002650}{ours} \\

Class. & {\cellcolor[gray]{\graystrength}\makecell
{\promptstyle{sheep_coco}{sheep} \\ \promptstyle{snowboard_coco}{snowboard}  }} 
& {\cellcolor[gray]{\graystrength}\makecell
{\promptstyle{donut_coco}{donut} \\ \promptstyle{couch_coco}{couch}  }}
& {\cellcolor[gray]{\graystrength}\makecell
{\promptstyle{tv/monitor}{tv/monitor} \\ \promptstyle{diningtable}{dining table} \\ \promptstyle{chair}{chair}
}} \\

\end{tabular}
\vspace{-8pt}
    \caption{
    \textbf{Visual ablations of the impact of background detection}. We show examples from COCO Object (left, middle) and PASCAL VOC (right). We note `bkg' our background refinement~\cref{sec:background}).
    }
    \label{fig:sup_bkg_abl}
\vspace{8pt}
\end{figure}

We now evaluate the quality of our background filtering using the class-agnostic foreground/background protocol defined in ~\cite{simeoni2023found}. We report in \cref{fig:sup:objdiscovery} the scores on the task of unsupervised object discovery (on VOC07~\cite{pascal-voc-2007}, VOC12~\cite{pascal-voc-2012} and COCO20k~\cite{lin2014microsoft} datasets with CorLoc metric) and unsupervised saliency detection in the `multi' setup of~\cite{simeoni2023found} (all results are provided when using post-processing bilateral solver on the classic DUT-OMRON~\cite{yang2013saliency}, DUTS-TEST~\cite{wang2017duts} and ECSSD~\cite{shi2015hierarchical} datasets, with the mIoU metric). For more details on the evaluation setup, we refer to~\cite{simeoni2023found}.
On both tasks, we observe on par or even better results than~\cite{simeoni2023found}, therefore showing the quality of our foreground predictions learnt from CLIP. 

\section{More qualitative results}

In this section, we illustrate the benefits of our method through additional comparative qualitative results.

\subsection{Visual ablations}

\parag{Our spatial pooling.} We show more examples of the application of our method \ours and compare it to MaskCLIP results in \cref{fig:sup_maskclip_abl}. We observe that in all cases, our pooling reduces the noise in the predictions and helps produce good-quality segmentation.

\parag{Our background filtering.}
By visualizing more results with and without the background refinement step in~\cref{fig:sup_bkg_abl}, we observe that the background refinement step helps remove uncertain segmentation such as the snow area (which was classified as `snowboard') in the left image, or on the cabinet, which is not annotated in VOC (right image). 

\subsection{In-the-wild examples}
\newlength{\cocowidth}
\newlength{\citywidth}
\setlength{\cocowidth}{0.14\textwidth}
\setlength{\citywidth}{0.3\textwidth}
\setlength{\resultsheight}{2.9cm}

\begin{figure*}[t]
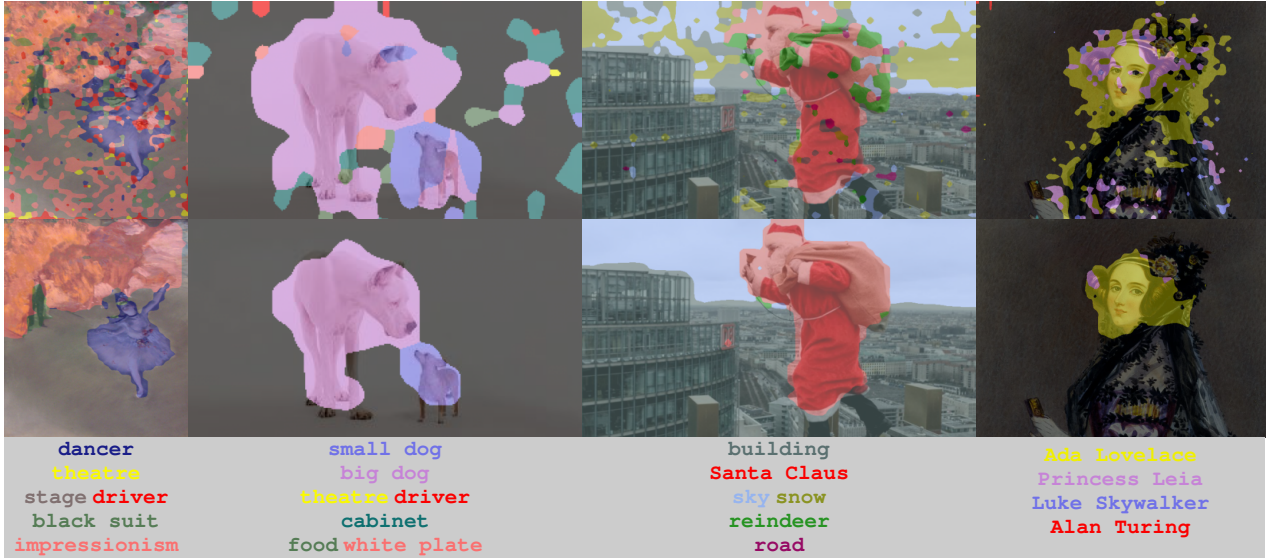

\centering
\renewcommand{\arraystretch}{0}
\setlength\tabcolsep{0pt} 

% \rowcolors{7}{lightgray}{white}
\begin{center}
\begin{tabular}{
>{\centering\arraybackslash}m{0.0em} 
>{\centering\arraybackslash}m{\cocowidth}
>{\centering\arraybackslash}m{\citywidth}
>{\centering\arraybackslash}m{\citywidth}
>{\centering\arraybackslash}m{0.22\textwidth}
}
 ~ & 
\addresult{\cocowidth}{degas}{maskclip} & 
\addresult{\citywidth}{dogsoriane}{maskclip} &
\addresult{\citywidth}{santa}{maskclip} &
\addresult{0.22\textwidth}{ada}{maskclip} 
\\
~ &
\addresult{\cocowidth}{degas}{ours} &
\addresult{\citywidth}{dogsoriane}{ours} &
\addresult{\citywidth}{santa}{ours} &
\addresult{0.22\textwidth}{ada}{ours} 
\\
\vspace{1pt}
 & 
 {\cellcolor[gray]{0.8} 
 \makecell{\promptstyle{dancer}{dancer} \\ \promptstyle{theatre}{theatre}  \\ \promptstyle{stage}{stage}  \promptstyle{driver}{driver} \\ 
  \promptstyle{black suit}{black suit}  \\
   \promptstyle{impressionism}{impressionism}}}
   &
{\cellcolor[gray]{0.8}
\makecell{\promptstyle{small dog}{small dog}  \\ \promptstyle{big dog}{big dog} \\ \promptstyle{theatre}{theatre} 
 \promptstyle{driver}{driver} \\ 
 \promptstyle{cabinet}{cabinet} \\ \promptstyle{food}{food} \promptstyle{white plate}{white plate} }}
 &
{\cellcolor[gray]{0.8}
\makecell{\promptstyle{building_wild}{building} \\ \promptstyle{driver}{Santa Claus} \\ \promptstyle{sky_wild}{sky} \promptstyle{snow}{snow} \\ \promptstyle{reindeer}{reindeer} \\ \promptstyle{road}{road}}}
 &
{\cellcolor[gray]{0.8}
\makecell{\promptstyle{ada}{Ada Lovelace} \\ \promptstyle{leia}{Princess Leia} \\ \promptstyle{luke}{Luke Skywalker} \\ \promptstyle{turing}{Alan Turing}}}
\\
\end{tabular}
\end{center}
    \vspace{-.5em}
    \caption{
    \textbf{In the wild comparative examples} between MaskCLIP (top) and \ours (bottom). While MaskCLIP generates noisy masks when prompted with \emph{false positive} classes our method is robust and produces cleaner masks. 
    }
    \label{fig:sup_wild}
\end{figure*}

 We show more in-the-wild examples in \cref{fig:sup_wild}, where we compare \ours against MaskCLIP. MaskCLIP produces very noisy masks, especially when multiple \emph{false positive} text queries are considered (we define such false positive queries as prompt queries that appear in the final segmentation but are not depicted in the image). Instead, \ours eliminates such false positive predictions and produces less noisy segmentation.

\subsection{Limitations}
\newlength{\pascalwidth}
\setlength{\pascalwidth}{0.16\textwidth}
\setlength{\cocowidth}{0.14\textwidth}
\setlength{\citywidth}{0.17\textwidth}
\setlength{\resultsheight}{1.8cm}

\begin{figure}[t]
\centering
\renewcommand{\arraystretch}{0}
\setlength\tabcolsep{0pt} 

\begin{center}
\begin{tabular}{
>{\centering\arraybackslash}m{0.0em} 
>{\centering\arraybackslash}m{\cocowidth}
>{\centering\arraybackslash}m{\citywidth}
>{\centering\arraybackslash}m{\pascalwidth}
% ccccccc
}
\rule{0pt}{2.5ex} &  
Object & City & ADE20K \\
 ~ & 
\addresultjpeg{\pascalwidth}{000000396568} & 
\addresult{\citywidth}{frankfurt_000001_038418}{} &
\addresultjpeg{\pascalwidth}{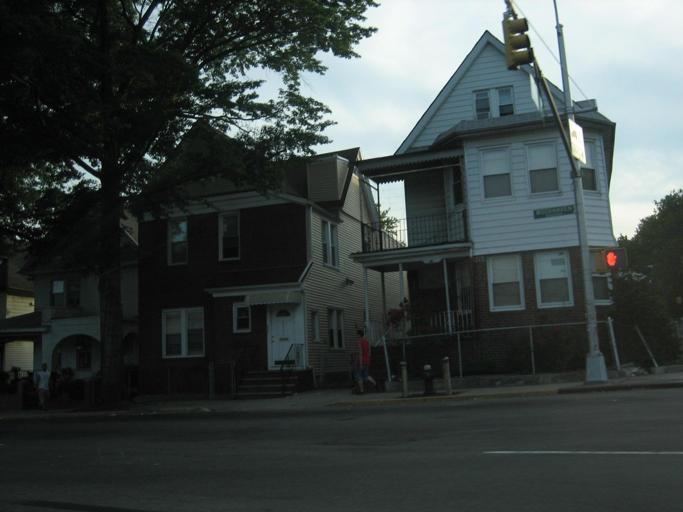} \\
~ &
\addresult{\pascalwidth}{000000396568}{gt} &
\addresult{\citywidth}{frankfurt_000001_038418}{gt} & 
\addresult{\pascalwidth}{ADE_val_00000856}{gt}
\\
~ & 
\addresult{\pascalwidth}{000000396568}{ours} &
\addresult{\citywidth}{frankfurt_000001_038418}{ours} & 
\addresult{\pascalwidth}{ADE_val_00000856}{ours} \\

\vspace{1pt}

 & 
 {\cellcolor[gray]{\graystrength} 
 \makecell{\promptstyle{train_coco}{train}}}
&
{\cellcolor[gray]{\graystrength}\makecell
{\promptstyle{traffic_sign_city}{traffic sign} \\ \promptstyle{road_city}{road} \promptstyle{vegetation_city}{vegetation} \\
 \promptstyle{car_city}{car} \promptstyle{sidewalk_city}{sidewalk} \\ \promptstyle{person_city}{person} \promptstyle{bicycle_city}{bicycle} }}
&
{\cellcolor[gray]{\graystrength}\makecell
{\promptstyle{house_ade}{house} \promptstyle{building_ade}{building} \\ 
\promptstyle{tree_ade}{tree} \promptstyle{sidewalk_ade}{sidewalk} \\
\promptstyle{person_ade}{person} \promptstyle{canopy_ade}{canopy} \\ \promptstyle{traffic light_ade}{traffic light}}} \\

\end{tabular}
\end{center}
    \vspace{-1.5em}
    \caption{
    \textbf{Failure cases} of our method. From top to bottom: input RGB image, ground truth (GT) masks, masks predicted by \ours, text prompts. We discuss these failure cases in \cref{ssec:failure_cases}.
    }
    \vspace{-0.5cm}
    \label{fig:sup_failures}
\end{figure}

\label{ssec:failure_cases}

We discuss here the known failure modes of our method \ours and visualize some in \cref{fig:sup_failures}.

We first observe some of the biases of CLIP, which for instance produces similar features for `train' and `train tracks' (left image), likely due to their frequent co-occurrence across images. 
We have observed other instances of this bias, 
e.g., for `boat' and `sea' queries. Second, although \ours can produce rather fine-grained segmentation (in terms of object sizes and classes), it can miss small or far-away objects as in Cityscapes (middle image). Finally, as with other open-vocabulary semantic segmentation methods, \ours is not robust to the ambiguities of the text queries. The example from ADE20K (right image) is such a case, where `house' is mistaken for `building'. In our experiments, we observed multiple segmentation ambiguities and we believe that the redefinition of evaluation metrics could help address the issue. We stress that the current evaluation setup, which is taken directly from fully supervised settings, might be limiting in an open-vocabulary paradigm.

\end{document}